\pdfoutput=1
\documentclass[11pt]{article}

\usepackage[final]{acl}
\usepackage{tabularx}
\usepackage{adjustbox}
\usepackage{booktabs}
\usepackage{times}
\usepackage{latexsym}
\usepackage{pifont}
\usepackage{listings}
\usepackage{array}
\usepackage{tikz}
\usepackage{listings}

\usepackage[T1]{fontenc}
\usepackage[utf8]{inputenc}

\usepackage{microtype}

\usepackage{inconsolata}

\usepackage{graphicx}
\usepackage{amsmath}
\usepackage{colortbl}
\usepackage{multirow}
\usepackage{xspace}

 \newcommand{\xhdr}[1]{\vspace{1.7mm}\noindent{{\bf #1.}}}

\newcommand{\benchname}{\textsc{BLADE\xspace}}

\newcommand{\numdatasets}{12\xspace}
\newcommand{\numdecisions}{536\xspace}
\newcommand{\nummcqs}{188\xspace}

\newcommand{\eqColValues}{\boldsymbol{\mathcal{S}}}

\newcommand{\eqAllTransforms}{\boldsymbol{\mathcal{T}}}

\newcommand{\eqAllGraphs}{\boldsymbol{\mathcal{G}}}

\newcommand{\eqCvars}{\boldsymbol{C}}
\newcommand{\eqCvarDesc}{c_{desc}}
\newcommand{\eqCvarType}{c_{type}}
\newcommand{\eqCvarCols}{C_{cols}}
\newcommand{\eqCvarColsi}{C_{cols,i}}

\newcommand{\eqModelDesc}{m_{desc}}
\newcommand{\eqModelCols}{M_{cols}}

\newcommand*\circled[1]{\tikz[baseline=(char.base)]{
            \node[shape=circle,draw,inner sep=2pt] (char) {#1};}}

\newcommand{\githubURL}{https://github.com/behavioral-data/BLADE}

\usepackage{graphicx}
\usepackage{wrapfig}

\title{\raisebox{-0.2\height}{\includegraphics[scale=0.015]{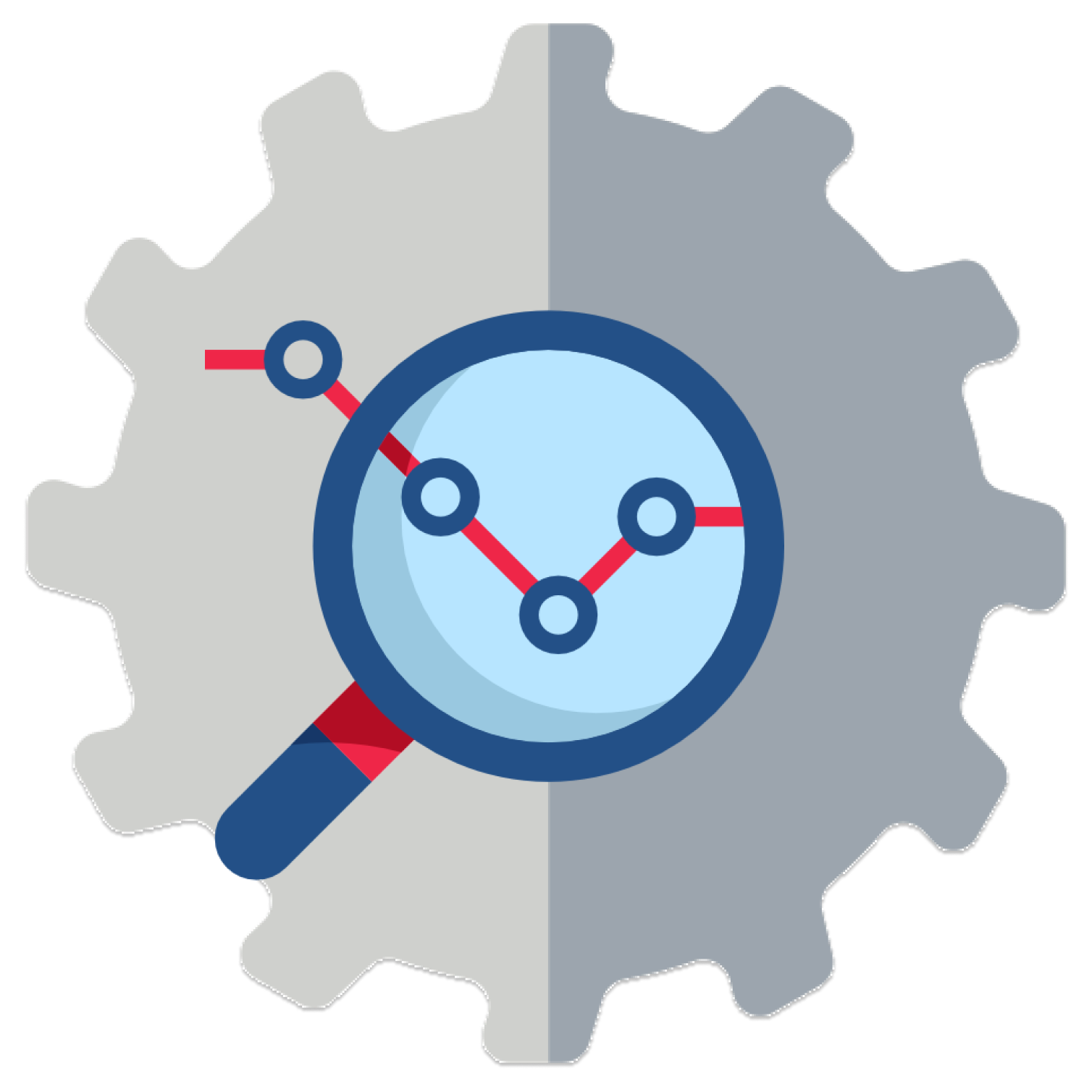}}~\benchname: Benchmarking Language Model Agents\\ for Data-Driven Science}

\author{
  \textbf{Ken Gu\textsuperscript{1}}
  Ruoxi Shang\textsuperscript{1}
  \textbf{Ruien Jiang\textsuperscript{2}}\footnotemark[1]
  Keying Kuang\textsuperscript{2}\footnotemark[1]
  \textbf{Richard-John Lin\textsuperscript{3}}\footnotemark[1]  \\
  \textbf{Donghe Lyu\textsuperscript{4}\footnotemark[1]}
\textbf{Yue Mao\textsuperscript{5}\footnotemark[1]}
  \textbf{Youran Pan\textsuperscript{3}\footnotemark[1]} 
  \textbf{Teng Wu\textsuperscript{6}}\footnotemark[1]
  \textbf{Jiaqian Yu\textsuperscript{7}}\footnotemark[1]
  \textbf{Yikun Zhang\textsuperscript{1}}\footnotemark[1] \\
  \textbf{Tianmai M. Zhang\textsuperscript{1}}\footnotemark[1] 
  \textbf{Lanyi Zhu\textsuperscript{1}}\thanks{These authors contributed equally to this work.}
  \textbf{Mike A. Merrill\textsuperscript{1}}
  \textbf{Jeffrey Heer\textsuperscript{1}}
  \textbf{Tim Althoff\textsuperscript{1}} \\ 
  \textsuperscript{1}University of Washington
  \textsuperscript{2}UC Berkeley 
  \textsuperscript{3}New York University
  \textsuperscript{4}Stanford University \\
  \textsuperscript{5}University of British Columbia
  \textsuperscript{6}Microsoft 
  \textsuperscript{7}George Washington University \\
  ~\textbf{\url{\githubURL}}
}

\usepackage{enumitem}

\begin{document}
\maketitle
\DeclareRobustCommand{\mychar}{%
  \begingroup\normalfont
  \includegraphics[height=\fontcharht\font`\B]{figs/logo.png}%
  \endgroup
}

\newcommand{\figMain}{
\begin{figure*}[t!]
    \centering
  \includegraphics[width=1.0\linewidth]{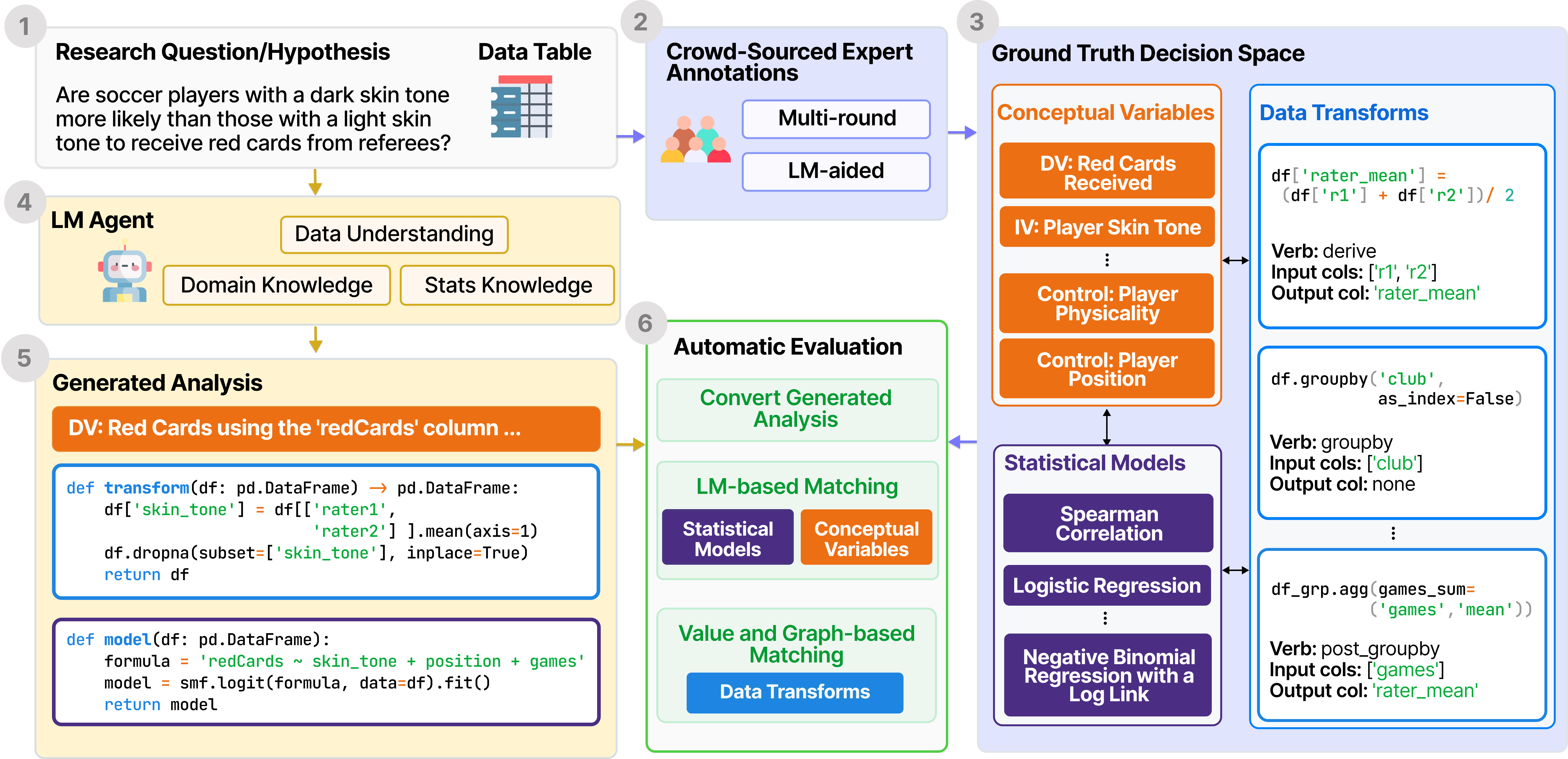}
  \caption{Overview of \benchname. We gathered research questions and datasets from existing research papers, crowd-sourced analysis studies and statistic textbooks as well as analyses from expert annotators (boxes 1-2-3, and Sec.~\ref{sec:benchmark_consturction}). Given a research question and dataset, LM agents generate a full analysis containing the relevant conceptual variables, a data transform function, and a statistical modeling function (boxes 1-4-5, and Sec.~\ref{sec:task_gen}). \benchname~automatically evaluates this against the ground truth (box 6 and Sec.~\ref{sec:enable_eval}).}
  %\sandy{Excellent figure! Where do you mention item 6? Please add, with a paper section reference.}
  \label{fig:main}
\end{figure*}
}

\newcommand{\figExample}{
\begin{figure*}[t!]
    \centering
  \includegraphics[width=0.72\linewidth]{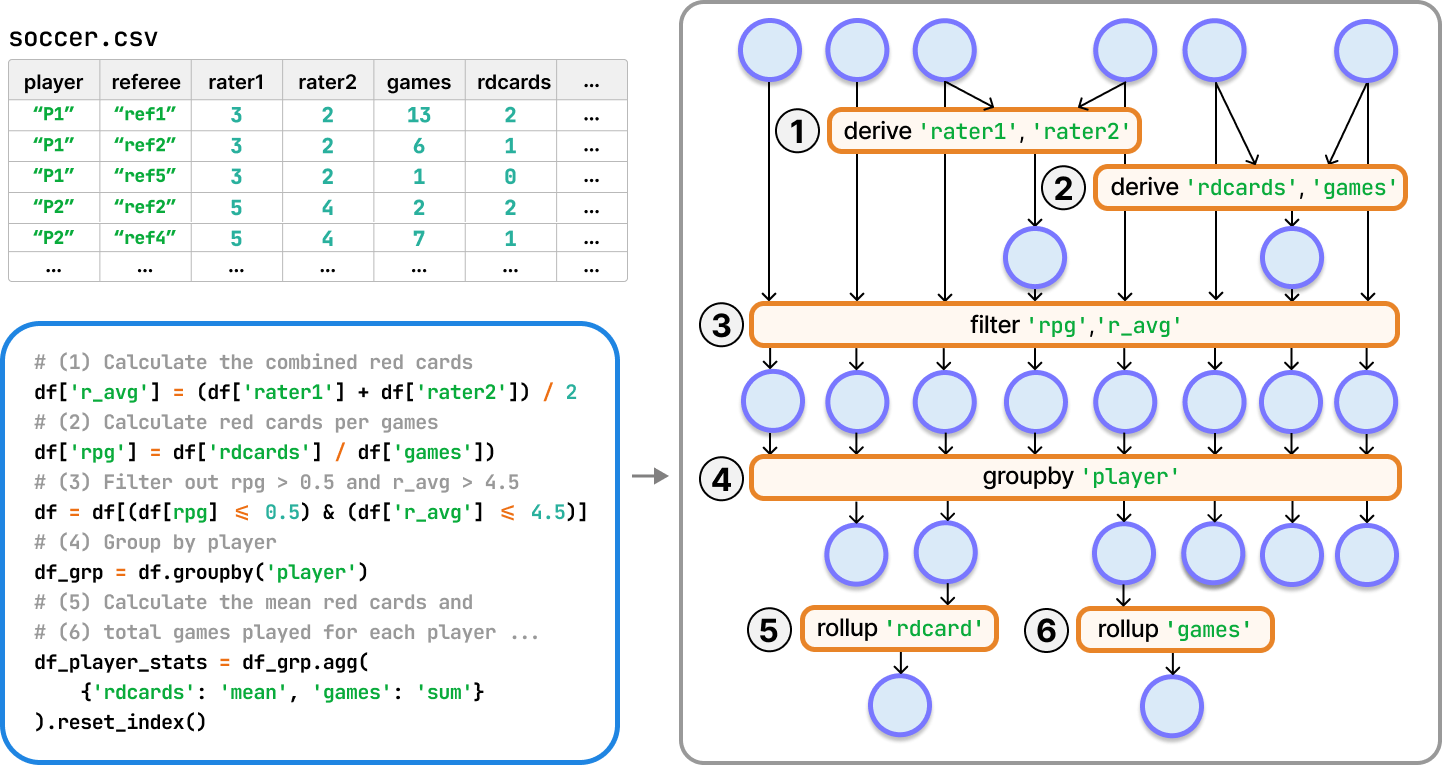}
  \caption{To allow flexible and fine-grained matching, we represent transforms in code (left) as a column data flow graph $G$ (right). The nodes in blue are column indicator nodes $P$, and the nodes in orange are transform nodes $T$. Details of the data flow graph formalization are in Appendix~\ref{appendix:matching_def}}.
  \label{fig:transform_example}
\end{figure*}
}

\newcommand{\figConvert}{
\begin{figure}[t!]
    \centering
  \includegraphics[width=0.9\linewidth]{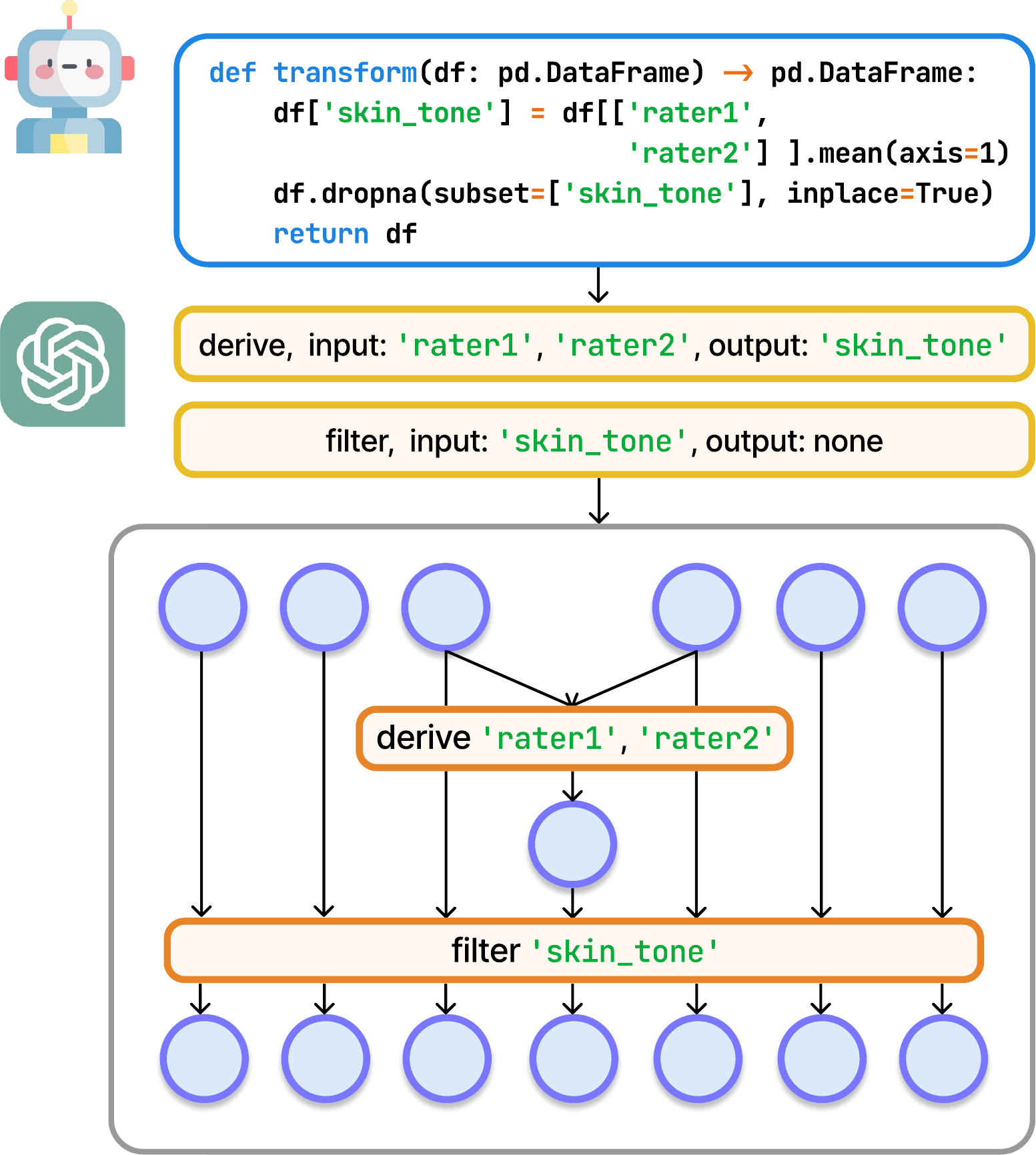}
  \caption{Given a transform function from a graph (top), we first use an LM (GPT-4o) to convert the transform into individual transform units with verb and column specifications (middle). Using this information, we then derive the column data flow graph $G$ (bottom).}
  \label{fig:convert}
\end{figure}
}

\newcommand{\figExampleSubmission}{
\begin{figure*}[t!]
    \centering
  \includegraphics[width=1.0\linewidth]{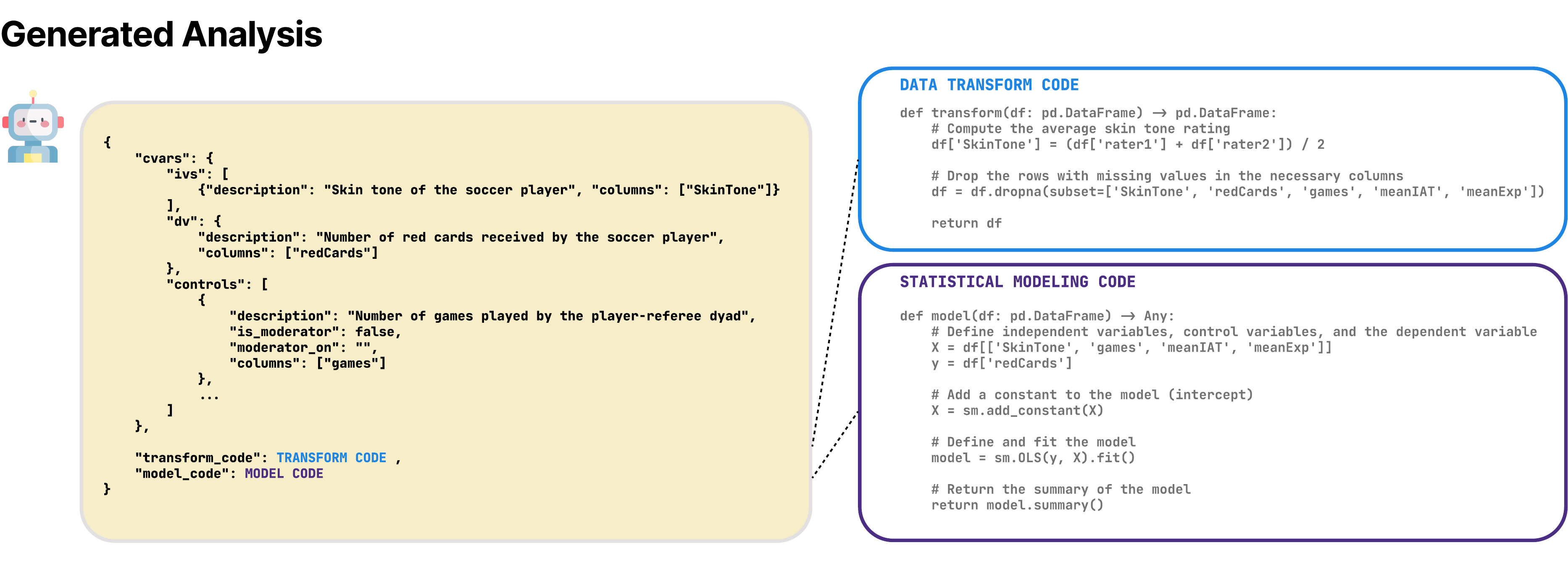}
  \caption{Example of the full analysis submission to \benchname.}
  \label{fig:example_submission}
\end{figure*}
}

\newcommand{\relatedBenchTable}{
\begin{table*}[t!]
\centering
\small
\begin{tabular}{l|cccccc}
\textbf{Requirements} &
  \tiny{\textbf{Data Interpreter}} &
  \tiny{\textbf{MLAgentBench}} &
  \tiny{\textbf{QRData}} &
  \tiny{\textbf{DS-Agent}} &
  \tiny{\textbf{DABench}} &
  {\textbf{Ours}} \\ 
 &
  \tiny{\cite{hong2024data}} &
  \tiny{\cite{huang2023benchmarking}} &
  \tiny{\cite{liu2024llms}} &
  \tiny{\cite{Guo2024DSAgentAD}} &
  \tiny{\cite{hu2024infiagent}} &
  \multicolumn{1}{l}{} \\ 
  \rowcolor[HTML]{F0F0F0} \textbf{Agent abilities tested} & & & & & &\\
(1) comprehend data semantics   & $-$ & $-$ & \textcolor{teal}{\ding{52}} & $-$ & $-$ & \textcolor{teal}{\ding{52}} \\
(2) integrate domain knowledge  & $-$ & $-$ & \textcolor{red}{\ding{55}}  & $-$ & $-$ & \textcolor{teal}{\ding{52}} \\
(3) conduct multi-step reasoning & \textcolor{teal}{\ding{52}} & \textcolor{teal}{\ding{52}} & $-$ & \textcolor{teal}{\ding{52}} & $-$ & \textcolor{teal}{\ding{52}} \\
(4) discern justifiable decisions      & \textcolor{red}{\ding{55}}  & \textcolor{red}{\ding{55}} & \textcolor{red}{\ding{55}} & \textcolor{red}{\ding{55}} & \textcolor{red}{\ding{55}} & \textcolor{teal}{\ding{52}} \\ \rowcolor[HTML]{F0F0F0} \textbf{Evaluation characteristics} & & & & & & \\
(5) automatic    & \textcolor{teal}{\ding{52}} & \textcolor{teal}{\ding{52}} & \textcolor{teal}{\ding{52}} & \textcolor{teal}{\ding{52}} & \textcolor{teal}{\ding{52}} & \textcolor{teal}{\ding{52}} \\ 
(6) decision-based  & \textcolor{red}{\ding{55}} & \textcolor{red}{\ding{55}}& \textcolor{red}{\ding{55}} & \textcolor{red}{\ding{55}} & \textcolor{red}{\ding{55}} & \textcolor{teal}{\ding{52}} \\
(7) flexible to decision input  &  \textcolor{red}{\ding{55}} &  \textcolor{red}{\ding{55}} &  \textcolor{red}{\ding{55}}  &  \textcolor{red}{\ding{55}}  &  \textcolor{red}{\ding{55}} & \textcolor{teal}{\ding{52}}
\end{tabular}
\caption{Comparing \benchname~against existing data analysis evaluation datasets and benchmarks for conducting~\textit{scientific analyses} based on the requirements specified in Section \ref{sec:benchmark_desiderata}. $-$ indicates partial satisfaction (e.g., data understanding is only on ML model building). See Table~\ref{tab:benchmark-comparison} for examples from \benchname~and recent benchmarks.}
\label{tab:main_compare}
\end{table*}
}

\newcommand{\datasetmetadata}{
\begin{table*}[h!]
%\begin{adjustbox}{angle=90}
\renewcommand{\arraystretch}{1.2}
\footnotesize
\begin{adjustbox}{max width=\textwidth} 
\begin{tabular}{p{1.8cm}p{1.8cm}p{3.5cm}p{3.5cm}p{2.9cm}}

\toprule
\textbf{Dataset} & \textbf{Domain} & \textbf{Keywords} & \textbf{Research Question} & \textbf{Source paper} \\ 
\midrule

hurricane &
  \begin{tabular}[t]{@{}m{1.8cm}@{}}\raggedright Behavioral \\ Sciences \end{tabular} &
  \begin{tabular}[t]{@{}m{3.5cm}@{}}\raggedright hurricane names, \\ gender stereotypes, \\ risk perception, \\ natural disasters \end{tabular} &
  \begin{tabular}[t]{@{}m{3.5cm}@{}}\raggedright Hurricanes with more feminine names are perceived as less threatening and hence lead to fewer precautionary measures by the general public. \end{tabular} &
  \begin{tabular}[t]{@{}m{2.9cm}@{}}\raggedright \cite{jung2014female, malter2014female, maley2014statistics,bakkensen2014population, simonsohn2020specification} \end{tabular} \\ \hline

mortgage &
  \begin{tabular}[t]{@{}m{1.8cm}@{}}\raggedright Finance and \\ Economics, \\ Demographics \end{tabular} &
  \begin{tabular}[t]{@{}m{3.5cm}@{}}\raggedright lending discrimination, \\ redlining, \\ credit risk, \\ fair housing \end{tabular} &
  \begin{tabular}[t]{@{}m{3.5cm}@{}}\raggedright How does gender affect whether banks approve an individual’s mortgage application? \end{tabular} &
  \begin{tabular}[t]{@{}m{2.9cm}@{}}\raggedright \cite{liu2020boba, munnell1996mortgage, young2017model} \end{tabular} \\ \hline

soccer &
  \begin{tabular}[t]{@{}m{1.8cm}@{}}\raggedright Behavioral\\ Sciences \end{tabular} &
  \begin{tabular}[t]{@{}m{3.5cm}@{}}\raggedright skin tone, \\ racial bias, \\ referee decisions, \\ sports analytics \end{tabular} &
  \begin{tabular}[t]{@{}m{3.5cm}@{}}\raggedright Are soccer players with a dark skin tone more likely than those with a light skin tone to receive red cards from referees? \end{tabular} &
  \begin{tabular}[t]{@{}m{2.9cm}@{}}\raggedright \cite{Silberzahn2018ManyAO, auspurg2021has} \end{tabular} \\ \hline

reading &
  \begin{tabular}[t]{@{}m{1.8cm}@{}}\raggedright Education \end{tabular} &
  \begin{tabular}[t]{@{}m{3.5cm}@{}}\raggedright dyslexia, \\ web accessibility, \\ reading comprehension, \\ user experience \end{tabular} &
  \begin{tabular}[t]{@{}m{3.5cm}@{}}\raggedright Does 'Reader View' -- a modified web page layout -- improve reading speed for individuals with dyslexia? \end{tabular} &
  \begin{tabular}[t]{@{}m{2.9cm}@{}}\raggedright \cite{li2019impact, liu2020boba} \end{tabular} \\ \hline

Fish &
  \begin{tabular}[t]{@{}m{1.8cm}@{}}\raggedright Health and \\ Well-being \end{tabular} &
  \begin{tabular}[t]{@{}m{3.5cm}@{}}\raggedright recreational fishing, \\ environmental conservation, \\ visitor demographics, \\ count data analysis \end{tabular} &
  \begin{tabular}[t]{@{}m{3.5cm}@{}}\raggedright How many fish on average do visitors take per hour, when fishing? \end{tabular} &
  \begin{tabular}[t]{@{}m{2.9cm}@{}}\raggedright \cite{mcelreath2018statistical} \end{tabular} \\ \hline

AMTL &
  \begin{tabular}[t]{@{}m{1.8cm}@{}}\raggedright Evolutionary \\ Biology \end{tabular} &
  \begin{tabular}[t]{@{}m{3.5cm}@{}}\raggedright antemortem tooth loss, \\ fossil hominins, \\ dental anthropology, \\ comparative anatomy \end{tabular} &
  \begin{tabular}[t]{@{}m{3.5cm}@{}}\raggedright Do modern humans have higher frequencies of antemortem tooth loss compared to non-human primate genera after accounting for the effects of age, sex, and tooth class? \end{tabular} &
  \begin{tabular}[t]{@{}m{2.9cm}@{}}\raggedright \cite{gilmore2013comparison, mcelreath2018statistical, konigsberg2013bayes} \end{tabular} \\ \hline

Boxes &
  \begin{tabular}[t]{@{}m{1.8cm}@{}}\raggedright Education, \\ Behavioral \\ Sciences \end{tabular} &
  \begin{tabular}[t]{@{}m{3.5cm}@{}}\raggedright cultural transmission, \\ social learning biases, \\ cognitive development, \\ cross-cultural research \end{tabular} &
  \begin{tabular}[t]{@{}m{3.5cm}@{}}\raggedright How do children's reliance on majority preference develop over growth in age across different cultural contexts? \end{tabular} &
  \begin{tabular}[t]{@{}m{2.9cm}@{}}\raggedright \cite{van2018development, mcelreath2018statistical} \end{tabular} \\ \hline

Crofoot &
  \begin{tabular}[t]{@{}m{1.8cm}@{}}\raggedright Evolutionary \\ Biology \end{tabular} &
  \begin{tabular}[t]{@{}m{3.5cm}@{}}\raggedright intergroup competition, \\ territorial behavior, \\ spatial analysis, \\ animal tracking \end{tabular} &
  \begin{tabular}[t]{@{}m{3.5cm}@{}}\raggedright How do relative group size and contest location influence the probability of a capuchin monkey group winning an intergroup contest? \end{tabular} &
  \begin{tabular}[t]{@{}m{2.9cm}@{}}\raggedright \cite{crofoot2008interaction, mcelreath2018statistical} \end{tabular} \\ \hline

Panda\_nuts &
  \begin{tabular}[t]{@{}m{1.8cm}@{}}\raggedright Evolutionary \\ Biology \end{tabular} &
  \begin{tabular}[t]{@{}m{3.5cm}@{}}\raggedright tool use, \\ skill acquisition, \\ social learning, \\ primate cognition \end{tabular} &
  \begin{tabular}[t]{@{}m{3.5cm}@{}}\raggedright How do age, sex, and receiving help from another chimpanzee influence the nut-cracking efficiency of western chimpanzees? \end{tabular} &
  \begin{tabular}[t]{@{}m{2.9cm}@{}}\raggedright \cite{boesch2019learning, mcelreath2018statistical} \end{tabular} \\ \hline

Affairs &
  \begin{tabular}[t]{@{}m{1.8cm}@{}}\raggedright Behavioral \\ Sciences, \\ Demographics \end{tabular} &
  \begin{tabular}[t]{@{}m{3.5cm}@{}}\raggedright infidelity, \\ marital satisfaction, \\ sexual behavior, \\ limited dependent variables \end{tabular} &
  \begin{tabular}[t]{@{}m{3.5cm}@{}}\raggedright Does having children decrease (if at all) the engagement in extramarital affairs? \end{tabular} &
  \begin{tabular}[t]{@{}m{2.9cm}@{}}\raggedright \cite{fair1978theory, kleiber2008applied, long2006regression} \end{tabular} \\ \hline
CASchools &
  Education &
  \begin{tabular}[t]{@{}m{3.5cm}@{}}\raggedright standardized testing, \\ school resources, \\ achievement gap, \\ education policy \end{tabular} &
  \begin{tabular}[t]{@{}m{3.5cm}@{}}\raggedright Is a lower student-teacher ratio associated with higher academic performance? \end{tabular} &
  \begin{tabular}[t]{@{}m{2.9cm}@{}}\raggedright \cite{kleiber2008applied, stock2020introduction} \end{tabular} \\ \hline

TeachingRatings &
  Education &
  \begin{tabular}[t]{@{}m{3.5cm}@{}}\raggedright student evaluations, \\ instructor characteristics, \\ gender bias, \\ higher education \end{tabular} &
  \begin{tabular}[t]{@{}m{3.5cm}@{}}\raggedright What is the impact of beauty on teaching evaluations received by teachers? \end{tabular} &
  \begin{tabular}[t]{@{}m{2.9cm}@{}}\raggedright \cite{simonsohn2020specification, hamermesh2005beauty, kleiber2008applied, stock2020introduction} \end{tabular} \\
  
   \bottomrule
\end{tabular}
\end{adjustbox}

\caption{Open-ended scientific research questions in \benchname~across different domains.}
\label{tab:benchmark_datasets}
\end{table*}
}

\newcommand{\taxonomyverbsPrev}{
\newpage
\begin{table*}[h!]
\small
\begin{adjustbox}{max width=\textwidth} % Adjust the table to fit the text width
\begin{tabular}{lllll}
\toprule
\textbf{Verb} &
  \textbf{Description} &
  \textbf{Input Columns} &
  \textbf{\begin{tabular}[c]{@{}l@{}}Affected \\ Output \\ Column(s)\end{tabular}} &
  \textbf{Example Code} \\ \midrule
Derive &
  \begin{tabular}[c]{@{}l@{}}Derive a new column value \\ based on the provided expressions.\end{tabular} & 
  Mandatory &
  One &
  \begin{tabular}[c]{@{}l@{}}\texttt{\# derive a new column 'sumXY' by}\\ \texttt{adding 'x' and 'y' df{[}'sumXY'{]} = df{[}'x'{]} + df{[}'y'{]}}\end{tabular} \\ \hline
  Filter &
  \begin{tabular}[c]{@{}l@{}}Filter a table to a subset of rows \\ based on the input criteria.\end{tabular} &
  Optional &
  All &
  \begin{tabular}[c]{@{}l@{}}\texttt{\# filter the dataframe to include only rows}\\ \texttt{where 'x' is greater than 2 df = df{[}df{[}'x'{]} \&amp;amp;gt; 2{]}}\end{tabular} \\ \hline
Slice & 
  \begin{tabular}[c]{@{}l@{}}Extract rows with indices from \\ start to end (end not included).\end{tabular} &
  Optional &
  All &
  \texttt{\# slice the dataframe to include rows 2 to 4 df = df.iloc{[}2:4{]}} \\ 
  \bottomrule
\end{tabular}
\end{adjustbox}
\caption{Taxonomy of transformation verbs used and captured in the ground truth analyses.}
\label{tab:transform_verbs}
\end{table*}
}

\newcommand{\taxonomyverbs}{
\begin{table*}[]
\renewcommand{\arraystretch}{1.3}
\small
\centering
\setlength{\tabcolsep}{6pt} % Adjust column spacing
\begin{tabular}{p{1.3cm} p{4cm} p{1.4cm} p{1.6cm} p{5cm}}
\toprule
\textbf{Verb} & \textbf{Description} & \textbf{Input Columns} & \textbf{Affected Output Column(s)} & \textbf{Example Code} \\ \midrule

Derive & Derive a new column value based on the provided expressions. & Mandatory & One &
\scriptsize\texttt{\# derive a new column 'sumXY' by adding 'x' and 'y'} 

\scriptsize\texttt{df{[}'sumXY'{]} = df{[}'x'{]} + df{[}'y'{]}} \\ \midrule

Filter & Filter a table to a subset of rows based on the input criteria. & Optional & All &
\scriptsize\texttt{\# filter the dataframe to include only rows where 'x' > 2} 

\scriptsize\texttt{df = df[df{[}'x'{]} > 2]} \\ \midrule

Slice & Extract rows with indices from start to end (end not included). & Optional & All &
\scriptsize\texttt{\# slice the dataframe to include rows 2 to 4} 

\scriptsize\texttt{df = df.iloc{[}2:4{]}} \\ \midrule

Groupby & Group table rows based on a set of column values. Groupby should return a pandas groupby object for subsequent operations. & Mandatory & All except groupby input columns &
\scriptsize\texttt{\# group the dataframe by 'x'} 

\scriptsize\texttt{grouped = df.groupby('x')} \\ \midrule

De-duplicate & De-duplicate table rows by removing repeated row values. & Optional & All &
\scriptsize\texttt{\# remove duplicate rows in the dataframe} 
\scriptsize\texttt{df = df.drop\_duplicates()} \\ \midrule

Impute & Impute missing values or rows. & Optional & One/All &
\scriptsize\texttt{\# replace NaN values with a specific value} 
\scriptsize\texttt{df = df.fillna(0)} \\ \midrule

Rollup & Rollup a table to produce an aggregate summary. This is used with groupby when aggregating a group. & Mandatory & One &
\scriptsize\texttt{df\_grp = df.groupby('x')} 

\scriptsize\texttt{\# rollup the grouped dataframe to get the mean of 'y'} 

\scriptsize\texttt{df = df\_grp.agg(mean\_y=('y', 'mean'))}
\scriptsize\texttt{\quad.reset\_index()} \\

\bottomrule
\end{tabular}
\caption{Taxonomy of transformation verbs utilized in the analysis ground truth. \benchname~leverages these verbs in its evaluation to measure the nuance and complexity inherent in transformation approaches (Appendix~\ref{appendix:match_transforms} explains our ``fuzzy'' transformation matching).}
\label{tab:transform_verbs}
\end{table*}
}

\newcommand{\annotatorinfo}{
\begin{table*}[]
\centering
\small
\renewcommand{\arraystretch}{1.2} % Adjust row spacing
\setlength{\tabcolsep}{8pt} % Adjust column spacing
\begin{tabular}{p{2cm} p{5cm} p{3.2cm} p{3cm}} 
\toprule
\textbf{Annotator ID} & \textbf{Current Occupation} & \textbf{Stats and Analysis Exp.} & \textbf{Analysis Frequency} \\ \midrule

A01  & PhD student in Statistics & 8 Years & A few times a week \\ 
A02  & PhD student in Statistics & 5 Years & A few times a week \\ 
A03  & PhD student in Statistics & 4 Years & A few times a week \\ 
A04  & PhD student in Biomedical and Health Informatics & 5 Years & A few times a week \\ 
A05  & PhD student in Measurement, Evaluation, and Research Methodology & 6 Years & A few times a week \\ 
A06  & Master's student in Communications & 5 Years & A few times a week \\ 
A07  & Master's student in Statistics & 6 Years & A few times a week \\ 
A08  & Data Scientist in the Finance Industry & 8 Years & Daily \\ 
A09  & Data Scientist in the Tech Industry & 8 Years & Daily \\ 
A10  & Data Scientist in the Tech Industry & 5 Years & Daily \\ 
A11  & Quantitative Researcher in Finance & 5 Years & Daily \\ 

\bottomrule
\end{tabular}
\caption{Expert level data annotation. All annotators have at least 4 years of experience in statistics and data analysis. In addition, they are either currently pursuing a postgraduate degree in a relevant scientific field or are regularly working with data in industry.}
\label{tab:annotator_background}
\end{table*}
}

\newcommand{\tabResultsNew}{

\begin{table}[h]
\centering
\small
\begin{tabular}{lc}
\hline
\multirow{1}{}{\textbf{Models}} & \textbf{F1  \hspace{0.65cm} (95\% CI)}  \\
 % & \multicolumn{1}{c}{} \ 
\hline
\multicolumn{2}{l}{\cellcolor[HTML]{EFEFEF}\textit{One-turn Setting}} \\
CodeLlama 7B & 16.8 \hspace{0.6cm} (15.2, 18.5)  \\
Deepseek-Coder 6.7B & 33.9 \hspace{0.6cm} (32.2, 35.4)  \\
Llama3 8B & 29.6 \hspace{0.6cm} (27.7, 31.5)  \\
Llama3 70B & 36.3 \hspace{0.6cm} (34.7, 37.8)  \\
Mixtral-8x22B & 40.1 \hspace{0.6cm} (38.0, 42.1)  \\
GPT-3.5 Turbo & 30.5 \hspace{0.6cm} (28.7, 32.2)  \\
GPT-4o & 41.7 \hspace{0.6cm} (40.2, 43.2) \\
Gemini 1.5 Pro & 41.1 \hspace{0.6cm} (39.6, 42.5) \\
Claude 3.5 Sonnet & \underline{43.9} \hspace{0.6cm} (42.6, 44.9)  \\
\hline
\multicolumn{2}{l}{\cellcolor[HTML]{EFEFEF}\textit{Agent Setting}} \\
Mixtral-8x22B & 40.8 \hspace{0.6cm} (38.2, 42.9)  \\
GPT-3.5 Turbo & 37.2 \hspace{0.6cm} (34.7, 39.7)  \\
GPT-4o & \underline{\textbf{44.8}} \hspace{0.6cm} (43.0, 46.3)  \\
Gemini 1.5 Pro & 40.1 \hspace{0.6cm} (38.3, 41.5)  \\
Claude 3.5 Sonnet & 43.1 \hspace{0.6cm} (41.4, 44.8)  \\
\hline
\end{tabular}
\caption{We report the decision-type weighted F1-score on analysis generation based on average precision and coverage@10. 
% See Appendix~\ref{appendix:f1_calc} for calculation details.
% where the weighting is applied to each decision type. }
}
\label{tab:results_main}
\end{table}
}

\newcommand{\tabResultsEmpty}{
\begin{table*}[t]
\begin{tabular}{lcc}
\hline
\multirow{2}{*}{\textbf{Models}} & \textbf{Average Hit Rate} & \textbf{Average Coverage} \\
 & \multicolumn{1}{c}{(95\% CI)} & \multicolumn{1}{c}{(95\% CI)} \\
\hline
\multicolumn{3}{l}{\cellcolor[HTML]{EFEFEF}\textit{One-turn Setting}} \\
l7b & & \\
d7b & & \\
llama3-8b & & \\
mixtral-8x22b & & \\
llama3-70b & & \\
gpt-3.5-turbo & & \\
gemini & & \\
gpt-4o-azure & & \\
claude-3.5-sonnet & & \\
\hline
\multicolumn{3}{l}{\cellcolor[HTML]{EFEFEF}\textit{Agent Setting}} \\
gemini\_agent & & \\
gpt-3.5-turbo\_agent & & \\
gpt-4o-azure\_agent & & \\
\hline
\end{tabular}
\caption{\textbf{Main results table.} We report average hit rate and coverage scores with 95\% confidence intervals across models, separated into one-turn and agent settings.}
\end{table*}
}

\newcommand{\tabResults}{
\begin{table*}[t]
\begin{tabular}{llllllllllll}
\hline
                                                     
                                                     & \multicolumn{5}{c}{\textbf{Average Hit Rate}}  & \multicolumn{5}{c}{\textbf{Average Coverage}} \\ \cline{2-6} \cline{7-11} 
\multirow{-2}{*}{\textbf{Models}} &
  \multicolumn{1}{r}{CV} &
  \multicolumn{1}{r}{T-V} &
  \multicolumn{1}{r}{T-G} &
  \multicolumn{1}{r}{M1} &
 \multicolumn{1}{r}{M2} &
  \multicolumn{1}{r}{CV} &
  \multicolumn{1}{r}{T-V} &
  \multicolumn{1}{r}{T-G} &
  \multicolumn{1}{r}{M1} &
  \multicolumn{1}{r}{M2} & \\
\multicolumn{12}{l}{\cellcolor[HTML]{EFEFEF}\textit{One-shot Setting}}                                                                            \\
CodeLlama-instruct &    54.1     &    36.9    &  11.5      &   59.6     &   17.9     &     59.2    &   15.7     &   15.9     &   41.8     &   5.5  &        \\
Deepseek-coder-instruct &   56.8      &   39.0     &   39.2     &   \textbf{84.1}     &   14.5     &   71.3      &  16.5   & 34.2       &   43.4     &  4.1    &        \\
GPT-3.5 Turbo &    79.9     &   39.2     &   33.0     &    56.3    &   \textbf{40.0 }    &   51.1      &  12.4       &   22.9     &   32.7     &  5.2    &        \\
Gemini 1.5 Pro &    \textbf{88.5}     &   54.7     &   \textbf{75.0}     &   76.7     &  35.4       &   69.8   &  13.2      &   31.6   &  41.6   &  5.0    &        \\
GPT-4o &    81.7     &   45.8     &  40.4      &   69.2     &  25.8      &    \textbf{80.2}    &    19.8    &   33.9     &   43.8     &   4.6   &        \\
\multicolumn{12}{l}{\cellcolor[HTML]{EFEFEF}\textit{Agent Setting}}                                                                               \\
GPT-3.5 Turbo     &     77.1    &   35.7     &   36.1     &   49.9     &  22.9      &    69.5     &   \textbf{21.5}     &   28.9     &  36.2      &  5.7    &        \\
Gemini 1.5 Pro       &    85.1          &    \textbf{57.0}     &    68.9    &   74.9     &    36.1    &    71.7    &   16.3      &    25.3   &   44.4     &    \textbf{6.7}    &                \\
GPT-4o                       &  76.3        &     48.9    &   53.5     &   73.3 &  24.4 &  79.2   &  20.0   & \textbf{44.5}   &   \textbf{51.9}    &  4.4       &               \\
\end{tabular}
\caption{\textbf{Main results table.} We report average coverage and hit rate scores across datasets in \benchname. CV: Conceptual Variables Matching, T-V: Transformations - Value Matching, T-G: Transformations - Fuzzy Graph Matching (see Appendix~\ref{appendix:match_transforms} for details on matching transforms). M1: Statistical Model - Semantic Matching, M2: Statistical Model - Conceptual Variable Matching. In calculating the results, we use the runs that did not end in "no generation." For transform-related matching, we also ignore runs that encountered execution errors.}
% (see Fig.~\ref{fig:result_type} for the error distribution).}
\end{table*} 
\label{tab:results_table}
}

\newcommand{\tabResultsPrev}{
\begin{table*}[t]
\small
\centering
\begin{tabular}{l|c|c|c|c|c|c|c|c|c|c}
\hline
& \multicolumn{5}{c|}{Metric group 1} & \multicolumn{5}{c}{Metric group 2} \
\cline{2-12}
    & 1 & 2 & 3 & 4 & 5 & 6 & 7 & 8 & 9 & 10 \\
\rowcolor[HTML]{EFEFEF} 
\multicolumn{11}{l}{One-shot models} \\
gpt-35-turbo-azure & & & & & & & & & & \\
gpt-4o-agent-bench-azure & & & & & & & & & & \\
gemini-1.5-pro-latest-gemini & & & & & & & & & & \\
meta-llama/CodeLlama-7b-Instruct-hf-huggingface & & & & & & & & & & \\
meta-llama/CodeLlama-13b-Instruct-hf-huggingface & & & & & & & & & & \\
deepseek-ai/deepseek-coder-6.7b-instruct-huggingface & & & & & & & & & & \\
\rowcolor[HTML]{EFEFEF}
\multicolumn{11}{l}{Agent models} \\
gpt-35-turbo-azure & & & & & & & & & & \\
gpt-4o-agent-bench-azure & & & & & & & & & & \\
gemini-1.5-pro-latest-gemini & & & & & & & & & & \\
meta-llama/CodeLlama-7b-Instruct-hf-huggingface & & & & & & & & & & \\
deepseek-ai/deepseek-coder-6.7b-instruct-huggingface & & & & & & & & & & \\
\hline
\end{tabular}
\caption{Performance of one-shot and agent-based language models against \benchname.}
\label{tab:results_table}
\end{table*}
}

\newcommand{\figResultsHumanEvalComparison}{
\begin{figure*}[h]
    \centering
  \includegraphics[width=0.99\linewidth]{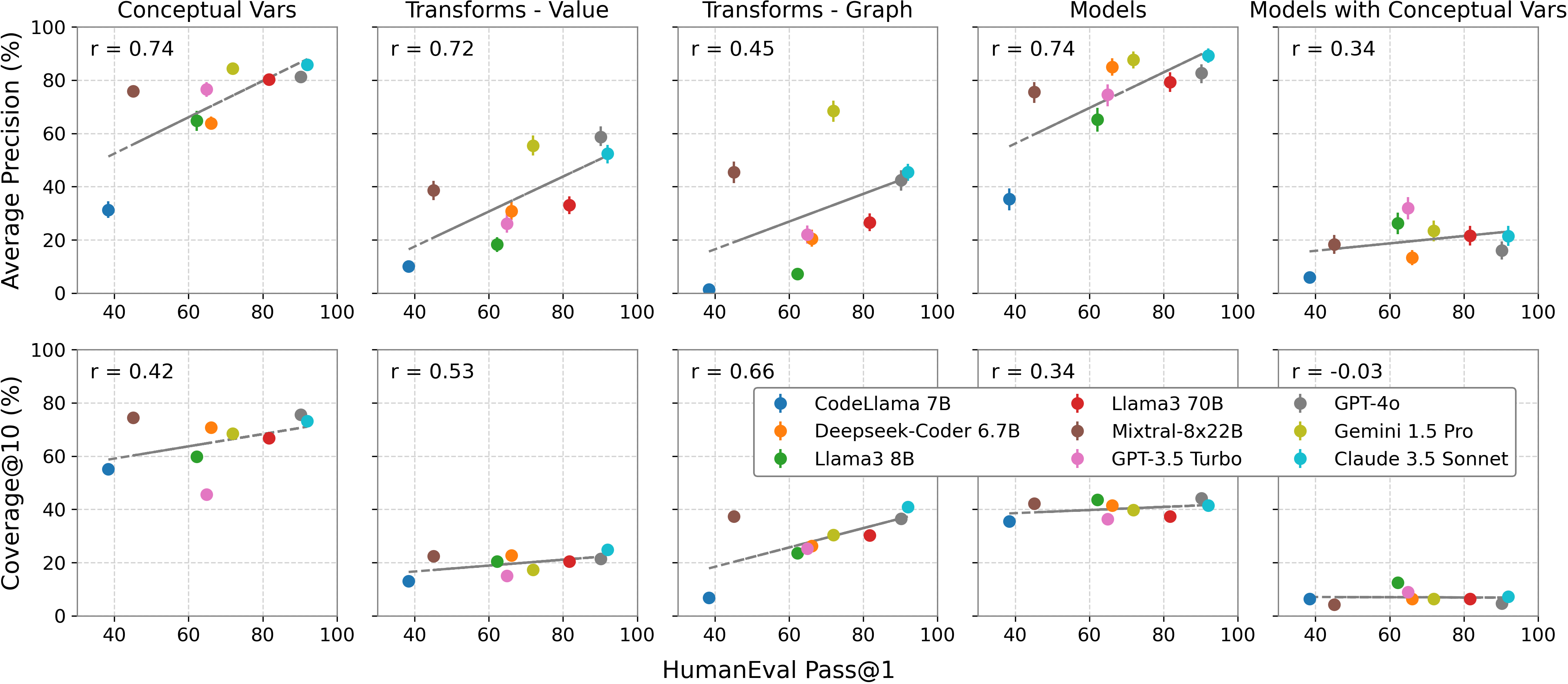}
  \caption{\benchname~Performance vs. HumanEval Performance. We compare BLADE evaluation metrics against reported Pass@1 on HumanEval~\cite{Chen2021EvaluatingLL} for all LMs in our experiments.}
  \label{fig:result_compare_humaneval}
\end{figure*}
}
\newcommand{\figResultsMain}{
\begin{figure*}[h]
    \centering
  \includegraphics[width=0.98\linewidth]{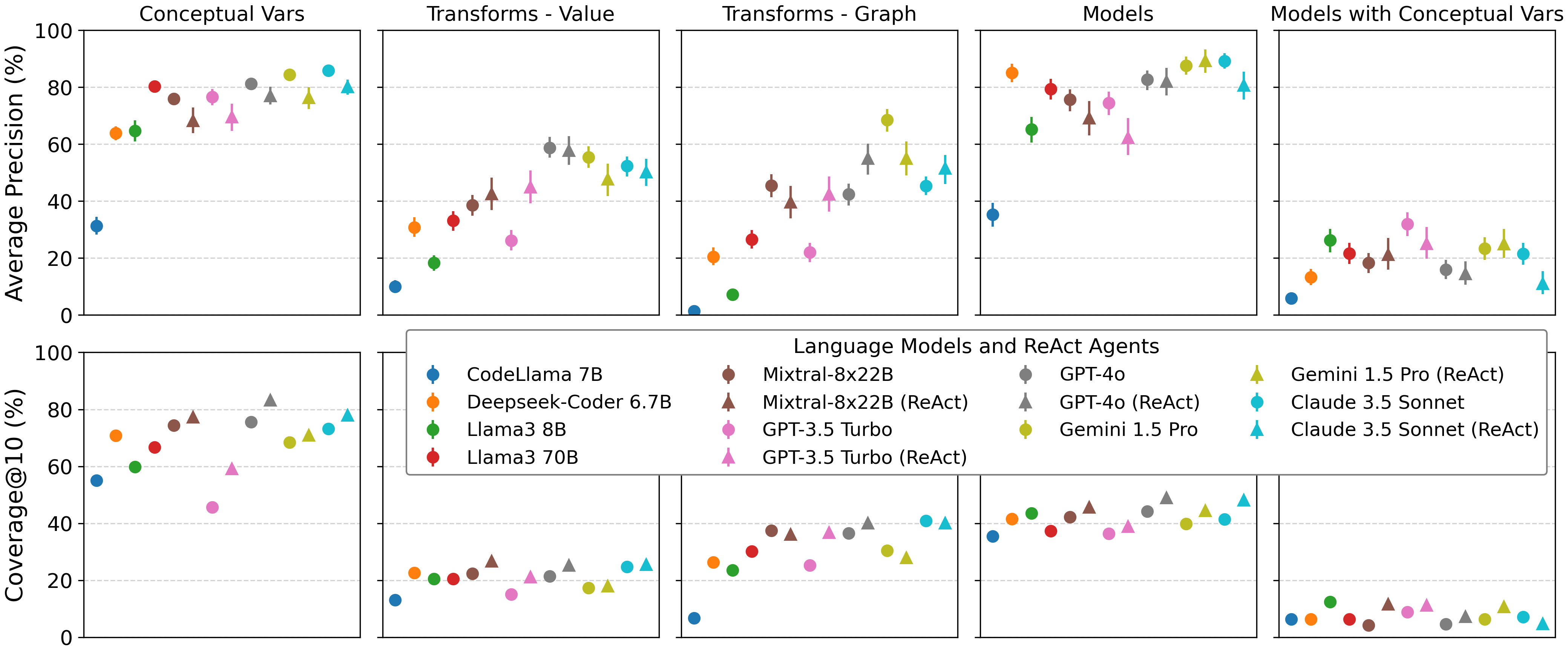}
  \caption{Average precision (top row) and coverage@10 (bottom row) percentages averaged across datasets in \benchname. All runs were included in the results. Run errors default to a hit rate of 0 and are counted in the coverage calculation (i.e., treated as a run that generated nothing). Error bars represent bootstrapped 95\% confidence intervals.}
  \label{fig:result_main}
\end{figure*}
}

\newcommand{\figResultsMCQ}{
\begin{figure}[h]
    \centering
  \includegraphics[width=0.95\linewidth]{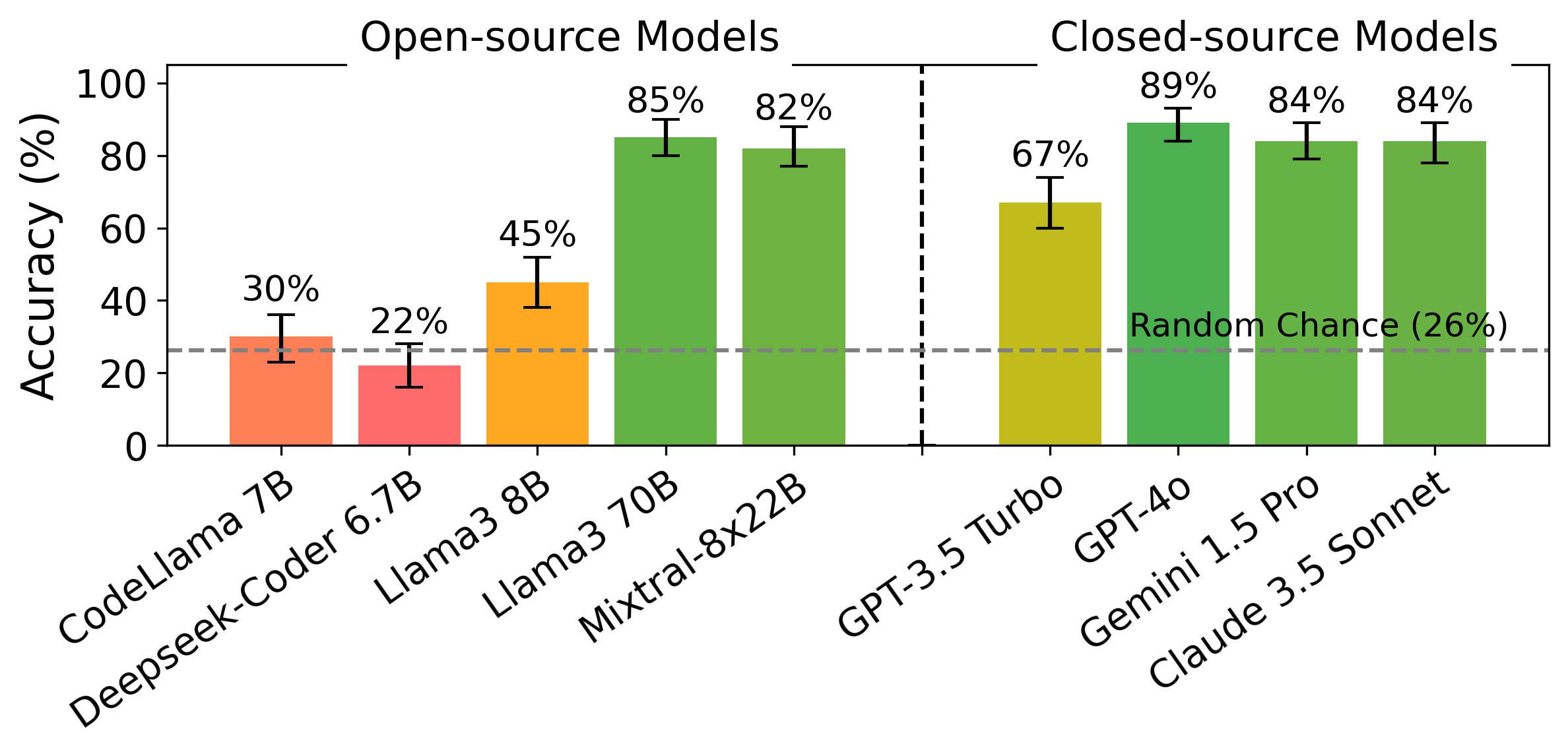}
  \caption{
  % MCQ results bar chart.
 Accuracy scores and 95\% confidence intervals for different models on \benchname's 188 MCQs (168 for transformations and 20 for conceptual variables).}
  \label{fig:mcq}
\end{figure}
}

\newcommand{\figResultsType}{
\begin{figure*}[h]
    \centering
  \includegraphics[width=0.91\linewidth]{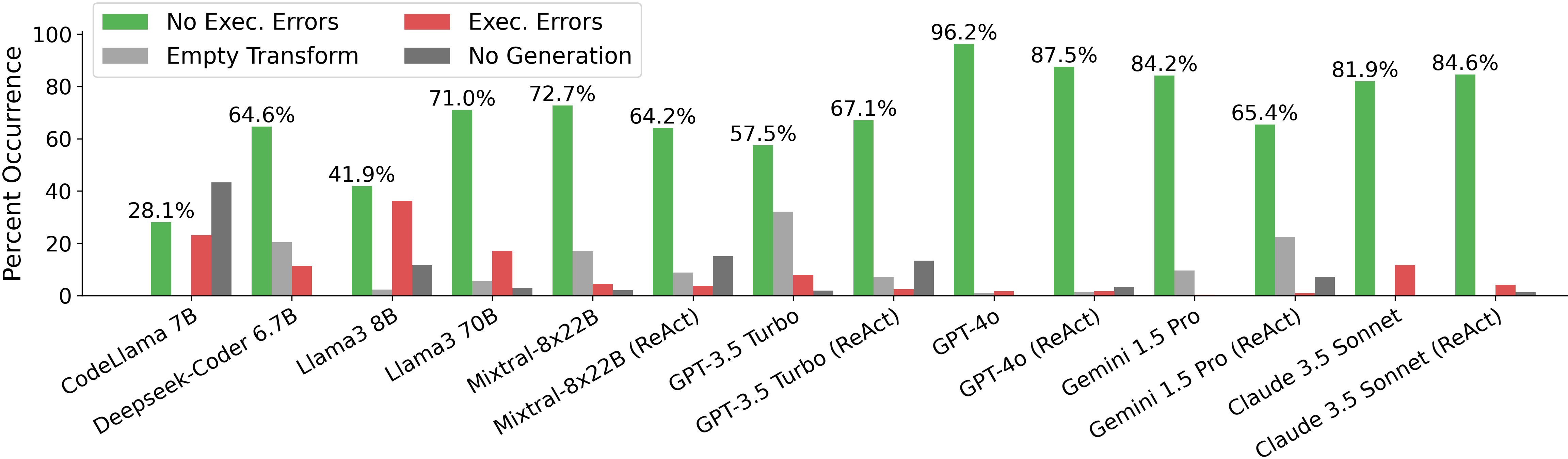}
  \caption{Characterization of run results for analysis generation for each LM and ReAct agent variants. "No execution errors" indicates executable transform code, "Empty transform" means no transformations were provided, "Execution errors" means the code resulted in errors, and "No generation" indicates the result could not be parsed.}
  \label{fig:result_type}
\end{figure*}
}

\newcommand{\tabDatasetSpecs}{
\begin{table*}[h]
\centering
\begin{tabular}{l|llllll}
 & Number of conceptual variables & Number of transformation & Number of model specs & Number of unique model types & Number of rows & Number of cols \\ \cline{1-6}
hurricane       & 4  & 6  & 18 & 3 & 94     & 14 \\
mortgage        & 12 & 23 & 6  & 2 & 2380   & 14 \\
soccer          & 16 & 77 & 41 & 9 & 146028 & 27 \\
reading         & 18 & 23 & 12 & 3 & 2568   & 20 \\
Fish            & 6  & 12 & 10 & 5 & 250    & 6  \\
AMTL            & 7  & 20 & 7  & 3 & 1450   & 9  \\
Boxes           & 5  & 13 & 5  & 3 & 629    & 5  \\
Crofoot         & 8  & 14 & 11 & 2 & 58     & 12 \\
Panda\_nuts     & 12 & 23 & 31 & 5 & 84     & 7  \\
Affairs         & 7  & 9  & 11 & 5 & 601    & 10 \\
CASchools       & 7  & 5  & 6  & 1 & 420    & 15 \\
TeachingRatings & 17 & 14 & 9  & 4 & 463    & 13
\end{tabular}
\end{table*}
}

\newcommand{\tabHorizontalComparison}{
\begin{table*}[t!]
\footnotesize
\centering
\setlength{\tabcolsep}{4pt}
\renewcommand{\arraystretch}{1.3}
\begin{adjustbox}{max width=\textwidth}
\begin{tabular}{|p{1.25cm}|p{4.45cm}|p{4.45cm}|p{4.45cm}|}
\hline
% & \textit{Example Task 1} & \textit{Example Task 2} & \textit{Example Task 3} \\
% \hline
\multicolumn{4}{|c|}{\textbf{BLADE}} \\
\hline
Question & 
Are soccer players with a dark skin tone more likely than those with a light skin tone to receive red cards from referees? &
How do age, sex, and receiving help from another chimpanzee influence the nut-cracking efficiency of western chimpanzees? &
Does 'Reader View' – a modified web page layout – improves reading speed for individuals with dyslexia? \\
\hline
Answer(s) &
\textit{16} conceptual variables, \textit{77} transformations, and \textit{41} modeling decisions &
\textit{11 }conceptual variables, \textit{19} transformations, and \textit{31} modeling decisions &
\textit{18} conceptual variables, \textit{24} transformations, and \textit{12} modeling decisions \\
\hline
\multicolumn{4}{|c|}{\textbf{ARCADE \cite{Yin2022NaturalLT}}} \\
\hline
Question &
How many male and female employees are born in 1992? &
Which countries host at least two Olympic games? &
What is the most expensive phone in each brand?\\
\hline
Answer(s) &
Two number counts &
A list of country names &
Dataframe of brand, model and price \\
\hline 
\multicolumn{4}{|c|}{\textbf{DABench \cite{hu2024infiagent}}} \\
\hline
Question &
Calculate the correlation coefficient between the "High Price" column and the "Low Price" column. &
Calculate the mean fare paid by the passengers. &
Categorize passengers into age groups and calculate mean fare for each group. \\
\hline
Answer(s) &
["relationship\_type", "linear"], \newline ["correlation\_coefficient", "0.99"] &
["mean\_fare", "34.65"] &
["mean\_fare\_elderly", "43.47"], \newline["mean\_fare\_teenager", "31.98"], \newline["mean\_fare\_child", "31.09"], \newline["mean\_fare\_adult", "35.17"] \\
\hline
\multicolumn{4}{|c|}{\textbf{MLAgentBench \cite{Huang2023MLAgentBenchEL}}} \\
\hline
Question &
\texttt{[CIFAR-10]} Given a training script on a dataset train.py, improve upon the current model performance (trained with current hyperparmeters in train.py). The training epochs should be within 10 to save time. &
\texttt{[Feedback]} Go through the data\_description.txt file to understand the data and all the features. You can summarize it in your research logs to keep track of what all you have to do. Then fill in the provided train.py script to train a model and iterate over different models or feature selections to get a better performance. & \texttt{[IMDB]}
Fill out train.py to 1) finetune DistilBERT on the IMDb dataset to determine whether a movie review is positive or negative and 2) save per class probabilities for test set examples to submission.csv. \\
\hline
Answer(s) &
Predictions for ML classification &
Predictions for ML regression  &
Predictions for ML classification \\
\hline

\multicolumn{4}{|c|}{\textbf{DS-Agent \cite{Guo2024DSAgentAD}}} \\
\hline
Question &
\texttt{[Airline Reviews]} You are solving this machine learning tasks of regression: 
The dataset presented here (Airline reviews) comprises customer feedback for British Airways. Here, we provide the text reviews. Your task is to predict the corresponding rating in the range of {1-10} given the reviews in the test set. The evaluation metric is root mean squared error (RMSE).&
\texttt{[Bitcoin Price Prediction]} You are solving this machine learning tasks of regression: 
The dataset presented here (the BTC News to Bitcoin Price dataset) comprises a series of BTC news title. Your task is to predict the bitcoin price based on the given BTC news title in the test set. The evaluation metric is root mean squared error (RMSE). &
\texttt{[BoolQ]} You are solving this machine learning tasks of classification: 
The dataset presented here (the BoolQ dataset) comprises a series of passage-question pairs. Given a passage and a question, your task is to identify whether the question can be inferred from the passage, with 0 as False and 1 as True. The evaluation metric is accuracy. \\
\hline
Answer(s) &
Predictions for ML regression  &
Predictions for ML regression   &
Predictions for ML classification \\
\hline

\multicolumn{4}{|c|}{\textbf{ML-Benchmark for Data Interpreter \cite{hong2024data}}} \\
\hline
Question &
\texttt{[Titanic]} This is a Titanic passenger survival dataset, and your goal is to predict passenger survival outcomes. The target column is Survived. Perform data analysis, data preprocessing, feature engineering, and modeling to predict the target. Report accuracy on the eval data. &
\texttt{[Santander Customer]} This is a customer’s financial dataset. Your goal is to predict which customers will
make a specific transaction in the future. The target column is the target. Perform data
analysis, data preprocessing, feature engineering, and modeling to predict the target.
Report AUC on the eval data. & \texttt{[Santander Value]} This is a medical dataset with over fifty anonymized health characteristics linked to three age-related conditions. Your goal is to predict whether a subject has or has not been diagnosed with one of these conditions.The target column is Class. 
Perform data analysis... the target. 
Report F1 Score on the eval data.  \\
\hline
Answer(s) &
Predictions for ML classification &
Predictions for ML classification &
Predictions for ML regression \\
\hline

\end{tabular}
\end{adjustbox}
\caption{Comparison of task examples in \benchname~and related benchmarks. \benchname~prioritizes open-ended scientific research questions rather than ML prediction tasks or data analysis code execution, focusing on the analysis approach and allowing for multiple valid solutions.}
\label{tab:benchmark-comparison}
\end{table*}

}
\begin{abstract}
Data-driven scientific discovery requires the iterative integration of scientific domain knowledge, statistical expertise, and an understanding of data semantics to make nuanced analytical decisions, e.g., about which variables, transformations, and statistical models to consider. LM-based agents equipped with planning, memory, and code execution capabilities have the potential to support data-driven science. However, evaluating agents on such open-ended tasks is challenging due to multiple valid approaches, partially correct steps, and different ways to express the same decisions.
To address these challenges, we present \benchname, a benchmark to automatically evaluate agents' multifaceted approaches to open-ended research questions. \benchname~consists of \numdatasets~datasets and research questions drawn from existing scientific literature, with ground truth collected from independent analyses by expert data scientists and researchers.  
To automatically evaluate agent responses, we developed corresponding computational methods to match different representations of analyses to this ground truth. Though language models possess considerable world knowledge, 
our evaluation shows that they are often limited to basic analyses. 
However, agents capable of interacting with the underlying data demonstrate improved, but still non-optimal, diversity in their analytical decision making.  Our work enables the evaluation of agents for data-driven science and provides researchers deeper insights into agents' analysis approaches. 

\end{abstract}

\section{Introduction}
\figMain

Scientific data continues to accumulate rapidly, driven by advancements in scientific instrumentation and the digitization of information. However, practicing \textit{data-driven science} (i.e., answering research questions from data) remains difficult, requiring rigorous methodologies, an understanding of data values and semantics, statistical and domain expertise, and critical thinking to validate hypotheses and draw meaningful and justifiable conclusions~\cite{Jun2021HypothesisFE, Breznau2022ObservingMR, Baker20161500SL, Aarts2015EstimatingTR}. 

Language model (LM)-based agents~\cite{Sumers2023CognitiveAF, Wu2023AutoGenEN, Wang2023ASO}, pre-trained on web-scale data and equipped with memory and tool usage capabilities~\cite{Schick2023ToolformerLM}, have the potential to conduct and support data-driven science. They can reason about and interact with heterogeneous data representing subjects, objects, and processes of study in the ``external'' world~\cite{Majumder2024DatadrivenDW}. However, to facilitate their progress, \textit{we need a reliable method to evaluate and measure their performance}. 

Recent benchmarks have enabled progress. However, they focus on either (1) data analysis execution with straightforward tasks containing a single, final, easily evaluated answer (e.g., \textit{Calculate the mean and standard deviation of the "Mar.2019" column}~\cite{Hu2024InfiAgentDABenchEA, Yin2022NaturalLT, Liu2024AreLC}) or (2) tasks for machine learning (ML) (e.g., \textit{improve the accuracy of an ML model}~\cite{Hong2024DataIA, Huang2023MLAgentBenchEL, Guo2024DSAgentAD}). 
For scientific analyses, these tasks require limited integration of external knowledge, limited understanding of data semantics, and limited grounding in external scientific knowledge. In addition, these benchmarks evaluate only on single metrics, such as ML model accuracy or completion rate. However, in the process of data-driven scientific discovery, the many intermediary decisions in a multi-step analysis are themselves critical to identify, meaningfully assess, and differentiate in order to improve agent performance.

Evaluating agent performance on open-ended data-driven analyses, especially automatically, poses specific challenges. First, the \textit{natural flexibility in making analysis decisions}~\cite{Gelman2014TheSC, Gelman2019TheGO, Simmons2011FalsePositiveP} makes it hard to establish a single ground truth that encompasses all justifiable choices. Second, the \textit{heterogeneity of decisions} (e.g., regarding hyperparameters of a statistical model, variables choices, high-level approaches, etc.) complicates efforts to decide on the representation and abstraction of meaningful decisions. Finally, given multiple valid decisions and approaches, determining the \textit{criteria and method to assess the correctness and soundness of the agent's analysis }is difficult to quantify.

In this work, we introduce \benchname, a benchmark for the principled evaluation of LM agents used for data-driven scientific analyses. Given a research question (e.g., \textit{``Are soccer players with a dark skin tone more likely than those with a light skin tone to receive red cards from referees?''}~\cite{Silberzahn2018ManyAO, auspurg2021has}) and a dataset, \benchname~evaluates agents' ability to integrate external scientific and statistical knowledge with 
an understanding of the data to conduct rigorously justifiable data analyses.

To build \benchname, we \textit{collected a set of actual research questions and datasets} (Fig.~\ref{fig:main}.1) from research papers, crowd-sourced analysis studies, and statistics textbooks~(Sec.~\ref{sec:benchmark_consturction}). Then, inspired by prior crowd-sourced analysis studies~\cite{Silberzahn2018ManyAO, Schweinsberg2021SameDD}, we \textit{recruited expert data analysts and collected high-quality data analyses} (Fig.~\ref{fig:main}.2) through a crowd-sourced analysis for each research question (i.e., multiple analysts independently performing a single analysis). To ensure our benchmark captured a broad variety of defensible analysis approaches, we \textit{asked analysts to validate alternative decisions} from their peers and LM-generated decisions seeded by analysts' own decisions. For this process, we also collected negative examples of \textit{"unjustifiable" decisions} to use when testing agents' ability to discern justifiable ones. We then combined all unique decisions to form the \textit{ground truth} (Fig.~\ref{fig:main}.3). 

Next, based on studies outlining decision steps in the data analysis process~\cite{Gu2023HowDD, Liu2019PathsEP, Liu2020UnderstandingTR, Jun2021HypothesisFE}, we formulated tasks. These tasks tested the discernment and formulation of \textit{analytical decisions that reflect multiple levels of abstraction}, ranging from executable code implementing data transformations to higher-level planning of conceptual variables requiring external scientific knowledge (Sec.~\ref{sec:benchmark_tasks}).

Finally, given our data and task, we developed \textit{representations} and matching criteria for different types of analysis decisions. We also developed corresponding \textit{computational methods to enable automatic evaluation of agent responses} (Sec.~\ref{sec:enable_eval}).
\relatedBenchTable

Overall, \benchname~contains \nummcqs~multiple choice and \numdecisions~ground truth analysis decisions encompassing multiple justifiable analysis approaches across \numdatasets~real-world datasets and research questions. To illustrate its utility and assess benchmark performance, we evaluate different LMs and a standard ReAct agent~\cite{yao2023react} that interacts with a sandbox notebook environment (Sec.~\ref{sec:experiments}). 

In our results (Sec.~\ref{sec:results}), we find most LMs are decent at discerning decisions and generating non-empty executable analyses. However, these analysis are basic and lack diversity. In particular, LM's coverage of the ground truth for forming statistical models with conceptual variables is below 13\%, and for operationalizing variables, it is below 
27\% (Fig.~\ref{fig:result_main}). The baseline ReAct agent shows a consistent improvement in coverage, though with plenty of room for improvement.

Our main contributions are: (1) a rigorously expert-annotated benchmark and the first of its kind to evaluate agents’ analytical decisions on open-ended scientific research questions; (2) an evaluation framework to automatically assess agent responses on fine-grained aspects of the analysis; and (3) results on various LMs and a ReAct agent indicating their current strengths and limitations.

Our work takes the first step in evaluating agents for open-ended data-driven scientific discovery, advancing our understanding of their capabilities to collect data, generate hypotheses, conduct analyses, and interpret results to form valid, justifiable scientific conclusions. To support further research and development, we open source our benchmark and evaluation framework\footnote{\url{\githubURL}}.

\section{Benchmark Requirements}
\label{sec:benchmark_desiderata}
Our benchmark evaluates agents on answering open-ended, data-driven scientific questions, advancing current efforts that execute analysis code based on precise single-answer instructions. Existing benchmarks are limited in their ability to assess agent decision-making during analysis and often do not capture the full scope of their approaches. Our benchmark addresses these limitations by focusing on the following key requirements (Table~\ref{tab:main_compare}).

We maintain that the ideal benchmark would evaluate an agent's abilities to (1) \textit{comprehend data semantics}, understanding the semantic relationships between variables and what the data represents relative to the external world, (2) \textit{integrate domain knowledge}, i.e., findings from related literature and an understanding of a ``world model'', (3) \textit{conduct multi-step reasoning} and planning at different levels of abstraction, i.e., high level planning vs. lower level code execution, given domain knowledge, an understanding of the data, and execution outputs, and (4) \textit{differentiate  justifiable decisions} with firm theoretical or statistical support~\cite{simonsohn2020specification} from unjustifiable ones.

Additionally, the benchmark evaluation should be
(5) \textit{automatic}, requiring no human intervention, (6) \textit{decision-based}, with the ground truth reflecting the intermediary decisions, and (7) \textit{flexible to decision input}, being aware of multiple ways to specify the same decision. Requirements 1 through 4 inform our \textit{data collection process} (Sec.~\ref{sec:benchmark_consturction})  and task formulation (Sec.~\ref{sec:benchmark_tasks}). Requirements 5 through 7 inform our \textit{evaluation procedure} (Sec.~\ref{sec:enable_eval}). Ultimately, we assess whether agents can plan, develop, and execute a justifiable analysis to answer a real-world research question.

\section{Benchmark Data Collection}
\label{sec:benchmark_consturction}

We now describe our data collection process for research questions (RQs), data, and ground-truth analyses.

\xhdr{RQs and Data} We selected scientific-grade datasets and RQs directly from scientific publications, particularly those studied in meta-analysis papers~\cite{Silberzahn2018ManyAO, simonsohn2020specification, young2017model} and reproduced in statistics textbooks~\cite{Mcelreath2020StatisticalR, kleiber2008applied}. We chose these sources because they provide a multitude of complex analyses, and relevant properties that make analyses non-trivial and revealing of statistical knowledge. Table \ref{tab:benchmark_datasets} summarizes these RQs, datasets, source papers, and meta-analysis papers. During this process, we ensured that the datasets were clearly documented and sufficiently   
complex to require non-trivial analyses, i.e., expert annotators would be required to distinguish defensible from indefensible decisions.

\xhdr{Annotation Process} To gather ground truth analyses and ensure the highest quality annotations, we followed a procedure similar to those used in previous crowd-sourced analysis studies~\cite{Silberzahn2018ManyAO, Schweinsberg2021SameDD}. We recruited 11 trained analysis experts and engaged them in a multi-stage process to ensure quality. Our experts had a self-reported average of 6 years of experience, with 6 pursuing or holding a Ph.D. in a scientific field. Since one of 
our key contributions is the corpus of ground truth analyses, we invited our expert annotators to be co-authors of this paper. See Appendix~\ref{appendix:recruitment} for details on  recruitment.

We gave each expert an RQ, dataset, and dataset description, including details of each column. For each dataset, experts independently conducted their analyses, recording all decisions they made. This naturally resulted in multiple analytical approaches.  To broaden the scope of possible strategies, we used an LM (GPT-4) to generate additional decisions (prompts shown in Fig.~\ref{fig:prompt1}).  

To ensure high data quality, experts validated and annotated each other's and LM-generated decisions as justified or unjustified. Agreement rates among expert annotators were relatively high: 75\% for transformations and 80\% for conceptual variables. In contrast, agreement on LM-generated decisions was much lower, at 27\% for transformations and 13\% for conceptual variables, highlighting low agent performance on decision generation. Many of these lower-agreement decisions were excluded from the final ground truth dataset. 

Finally, we brought the team of experts together to discuss their decisions, resolve ambiguities, and establish consensus. Our ground truth thus reflects alternative approaches validated by multiple experts. See Appendix~\ref{appendix:annotation_training} for details of our annotation process.

\section{Benchmark Tasks}
\label{sec:benchmark_tasks}
We want the benchmark tasks to represent decisions that are vital to the analysis and to evaluate the key skills needed to conduct data-driven science (i.e., requirements 1-4 in Section~\ref{sec:benchmark_desiderata}). We draw from prior studying the scientific analysis process~\cite{Gu2023HowDA, Liu2019PathsEP, Liu2020UnderstandingTR} and focus on agents' ability to \textit{discern} (Sec.~\ref{sec:tasks_tested_discern}) and \textit{make} (Sec.~\ref{sec:task_gen})~\textit{planning} decisions, i.e., those requiring a process of reasoning about and then synthesizing the data, scientific domain, and statistical knowledge. Specifically we test the following decisions:

\vspace{1.0mm}\noindent(1) \textit{Formulating Conceptual Variables.} Agents should recognize independent variables (IVs) , dependent variables (DVs), and control variables based on domain knowledge and multi-step reasoning (requirements 2 and 3), e.g., ``Prior literature suggests player physicality influences the referee’s perception. We can consider physicality a control.''

\vspace{1.0mm}\noindent(2) \textit{Executing Data Transformations.} Agents should select relevant columns and apply transformations to operationalize conceptual variables, e.g., using BMI as a proxy for player physicality via \textit{``weight''} and \textit{``height''} columns.

\vspace{1.0mm}\noindent(3) 
 \textit{Implementing Statistical Models.} Agents should choose the appropriate statistical model based on conceptual variables and transformed data to address the research question, requiring in-depth knowledge of statistical methods and the underlying hypothesis~\cite{Jun2019TeaAH, Jun2021HypothesisFE}.

\figExample

\subsection{Task 1: Discern Justifiable Decisions}
\label{sec:tasks_tested_discern}

To evaluate how well agents can discern justifiable decisions (requirement 4), \benchname~includes the following \textbf{M}ultiple \textbf{C}hoice \textbf{Q}uestions. (MCQ1) Given the research question and the dataset, which conceptual variable is the \textit{most/least} justifiable for the analysis? 
(MCQ2) Given the research question, dataset, and conceptual variable of interest, which transformation is the \textit{most/least} justifiable to operationalize the variable?

Each multiple choice question includes one correct and one or more incorrect answers. Justifiable and unjustifiable decisions were gathered during expert reviews of each other's and LM-generated decisions. A decision was deemed justifiable if all experts agreed and unjustifiable if the majority considered it unjustifiable. In addition, for MCQ2, additional negative samples were gathered from transformations 
that were used to derive conceptual variable that differed from the one in the question (i.e., easier negative examples). In total, \benchname~contains \nummcqs~multiple choice questions.

\subsection{Task 2: Generate an End-to-end Analysis}
\label{sec:task_gen}
For this significantly more complex task, agents need to generate a complete end-to-end analysis given a research question and a dataset. Specifically, to test agent performance on key analysis decisions, agents are to submit the following artifacts (e.g., in Fig.~\ref{fig:main}.5 and~\ref{fig:example_submission}), each mapping to one type of decision.

\begin{enumerate}
    \item \textit{A list of conceptual variables}, each with a natural language description (e.g, player physicality), the variable type (i.e., an independent, dependent, or control variable), and the name of the column in the final transformed data table used in the statistical model.
    \item \textit{An executable transformation function}, which is given a data table as input and returns a data table after performing the transformations to operationalize the conceptual variables.
    \item \textit{A statistical model function}, which takes as input the transformed data table and returns the specified statistical model. 
\end{enumerate}

\section{Flexible Automatic Evaluation}
\label{sec:enable_eval}

To quantitatively measure the quality of agent-generated analyses (i.e., the agent-generated artifacts) in a way that is \textit{automatic}, \textit{decision-based}, and \textit{flexible to decision input} (requirements 5-7 in Sec.~\ref{sec:benchmark_desiderata}), we need concrete representations of analysis decisions and associated matching criteria. We now discuss the representation and matching procedure for each artifact in an agent's submission.

\xhdr{Matching Conceptual Variables} Because conceptual variables capture high-level constructs, two similarly specified constructs (i.e, \textit{player physicality} and \textit{how physically imposing the player is}) should have the same meaning as long as they have the same variable type (i.e., IV, DV, or Control). To match these specifications, we employed an LM (GPT-4o) to determine the semantic equivalence between two conceptual variables. We followed a procedure similar to ~\cite{Liang2023CanLL} which was validated on semantically matching academic reviews (prompt in Fig.~\ref{fig:prompt5}). Appendix~\ref{appendix:cv_match} contains further details.

\xhdr{Matching Data Transformations} Since there are many ways to express data analyses in code, even ways that could be perfectly equivalent, we require a representation that maps equivalent transformations to a single representation (i.e., requirement 7 -- \textit{flexible to decision input}). Taking the code in Figure~\ref{fig:transform_example} as an example, the ordering of the transforms \circled{1}-\circled{2}-\circled{3} or \circled{2}-\circled{1}-\circled{3} are functionally equivalent with respect to the final product of the computation (i.e., as long as \circled{1} and \circled{2} come before \circled{3}). In addition, transformations that result in the exact same output column values (with a small margin of error for floating point)  should be considered equivalent transformations. Likewise, getting a certain column's values correct should mean that all relevant prior steps were correct 
and that decisions for each relevant prior transformation were correct. 
For example, if a submission missed \circled{6} but still correctly calculated the ``rdcards'' after the groupby, then the agent still correctly performed steps \circled{1}-\circled{5}, deserving significant partial credit. In complex tasks such as scientific data analysis, such partial credit enables meaningful differentiation of model performance and progress. 

To capture the aforementioned nuances, we developed a representation for data transformations using a data flow graph~\cite{Kavi1986AFD} (Fig.~\ref{fig:transform_example}~right). These graphs are useful because any series of transformations in the order of a topological sort~\cite{Manber1989IntroductionTA} on the graph leads to the same result. In addition, our graph captures data flow at the column-level (i.e., all cell values in a single column) to enable subsequent matching at the granularity of columns. In doing so, we allow for matching on transforms that require and affect only a subset of columns in a data table (e.g., in~Fig.~\ref{fig:transform_example}, getting \circled{1} correct is independent of getting \circled{2} correct). Appendix~\ref{appendix:matching_def} describes our data flow graph formalism in greater detail. 

In addition, the transforms (i.e., orange nodes in Fig.~\ref{fig:transform_example}) in the data flow graph represent a discrete data transformation decision that was made in wrangling the data (requirement 6 -- \textit{decision-based} evaluation). Specifically, each transform is defined by a fixed set of transform verbs (Table~\ref{tab:transform_verbs}) that are based on existing data wrangling libraries (i.e., Arquero\footnote{\url{https://idl.uw.edu/arquero/}} and Vega \cite{satyanarayan2016vega}), expandable, and validated to cover every analysis decision in our benchmark.
To match transforms in \benchname, we applied an LM (GPT-4o) to convert the transformation function in an agent's submission to the individual transform units (prompt in Fig.~\ref{fig:prompt3_1} and ~\ref{fig:prompt3_2}). 
We then constructed the agent's transformation data flow graph and matched it with the ground truth. We match based on both the \textit{ column values } that are the output of any discrete transformation and a \textit{fuzzier graph isomorphism matching} that determines whether approximately the same steps were applied. Appendix~\ref{appendix:match_transforms} describes the matching procedures in detail.

\xhdr{Matching Statistical Models}
The implementation of statistical models and relevant parameters could be evaluated in multiple ways (i.e., code, natural language, or mathematical formulas ~\cite{Jun2021HypothesisFE, mcelreath2018statistical}). To prioritize the underexplored \textit{planning} aspects of statistical modeling~\cite{Gu2023HowDA}, we focus on being able to select the right model and conceptual variables. In principle, this representation could be extended to include code, hyperparameters, and more. We first used an LM (GPT-4o) prompt (Fig.~\ref{fig:prompt5a}) to convert the modeling function into a natural language specification of the model and the columns in the transformed data table that it used. Next, using another LM (GPT-4o) prompt (Fig.~\ref{fig:prompt6}), we compared this output with the ground truth natural language specifications of the model and associated conceptual variables based on semantic equivalence. See Appendix~\ref{appendix:model_match} for additional details.

\xhdr{Evaluation of LM Evaluation Modules}
To validate our LM-based evaluation modules, two authors independently reviewed a sample of 615 LM-generated outputs across multiple datasets. After an initial round of review and resolution of any disagreements, the modules achieved the following correctness rates:
93\% for matching conceptual variables, 97\% for translating transform code into transform units, 97\% for converting modeling code into a natural language specification, and 92\% for matching statistical models. These results were deemed sufficient for our evaluation purposes.

\section{Experiments}
\label{sec:experiments}
To establish a baseline and evaluate the performance of LM-based agents on \benchname, we selected the following models: GPT-3.5 Turbo, GPT-4o~\cite{openai-gpt4}, Gemini 1.5 Pro~\cite{google-gemini}, and Claude 3.5 Sonnet~\cite{claude-sonnet} to represent closed-source general-purpose LMs; Llama3 8B, Llama3 70B~\cite{llama3}, and Mixtral-8x22B~\cite{mixtral} for open-source LMs; and CodeLlama Instruct 7B~\cite{Rozire2023CodeLO} and DeepSeek-Coder Instruct 6.7B~\cite{Guo2024DeepSeekCoderWT} for coding-specific LMs. 

% \tabResults

\xhdr{Experiment Settings}
For the multiple choice questions (Sec.~\ref{sec:tasks_tested_discern}, Task 1) we evaluate each LM with a temperature of 0. To generate an end-to-end analysis (Sec.~\ref{sec:task_gen}, Task 2), we evaluate LMs in one turn with a one-shot example (prompt in Fig.~\ref{fig:prompt4}). In addition, we develop an agent (also with an example demonstration), based on the ReAct framework~\cite{yao2023react}, that interacts with a computational notebook environment containing the data, reflects on observations from executing the code, and generates next-step actions. 
We evaluate the ReAct agent on Mixtral-8x22b, GPT-3.5 Turbo, GPT-4o, Gemini 1.5 Pro, and Claude 3.5 Sonnet. Appendix~\ref{appendix:ReAct} contains additional details on the setup of the agent and the choice of LMs.

For each LM and setting (i.e., one turn vs. ReAct agent), to encourage diversity we set the temperature to 0.8 and record a total of 40 runs for the one-turn setting and 20 runs for the agent setting to consider for computational budget.  For all LMs used to facilitate the evaluation (i.e., conversion and semantic matching), we use GPT-4o with a temperature of 0. Appendix~\ref{appendix:propmts} includes all prompts for the baselines and LM-aided evaluation.

\xhdr{Evaluation Metrics}
For the multiple choice tasks (Task 1), we measure agents on \textit{accuracy}. For the generation tasks (Task 2), to measure an agent's ability to both generate justifiable analyses and capture the breadth of justifiable approaches, we calculate an adapted \textit{F1-score} for each type of analysis decision (conceptual variables, transformation, and statistical model). The F1-score takes the harmonic mean of \textit{average precision} across runs and \textit{coverage@k}. The former quantifies how well an agent’s response matched with the ground truth while the latter evaluates how comprehensive agents are in generating justifiable alternative analyses. In our experiments, we report average precision across \textit{all} runs and coverage for $k=10$ runs. Appendix~\ref{appendix:f1_calc} contains the full details of our evaluation metrics.

\section{Results}
\label{sec:results}

We report the performance of LMs on MCQs (Task 1) in Figure~\ref{fig:mcq} and the results of LMs and our ReAct agent for analysis generation (Task 2) in Table~\ref{tab:results_main} and Figure~\ref{fig:result_main}. Here, we summarize our main findings.
\figResultsMCQ

\figResultsMain
\begin{table}[h]
\centering
\small
\begin{tabular}{lc}
\hline
\multirow{1}{*}{\textbf{Models}} & \textbf{F1  \hspace{0.65cm} (95\% CI)}  \\
\hline
\multicolumn{2}{l}{\cellcolor[HTML]{EFEFEF}\textit{One-turn Setting}} \\
CodeLlama 7B & 16.8 \hspace{0.6cm} (15.2, 18.5)  \\
Deepseek-Coder 6.7B & 33.9 \hspace{0.6cm} (32.2, 35.4)  \\
Llama3 8B & 29.6 \hspace{0.6cm} (27.7, 31.5)  \\
Llama3 70B & 36.3 \hspace{0.6cm} (34.7, 37.8)  \\
Mixtral-8x22B & 40.1 \hspace{0.6cm} (38.0, 42.1)  \\
GPT-3.5 Turbo & 30.5 \hspace{0.6cm} (28.7, 32.2)  \\
GPT-4o & 41.7 \hspace{0.6cm} (40.2, 43.2) \\
Gemini 1.5 Pro & 41.1 \hspace{0.6cm} (39.6, 42.5) \\
Claude 3.5 Sonnet & \underline{43.9} \hspace{0.6cm} (42.6, 44.9)  \\
\hline
\multicolumn{2}{l}{\cellcolor[HTML]{EFEFEF}\textit{Agent Setting}} \\
Mixtral-8x22B & 40.8 \hspace{0.6cm} (38.2, 42.9)  \\
GPT-3.5 Turbo & 37.2 \hspace{0.6cm} (34.7, 39.7)  \\
GPT-4o & \underline{\textbf{44.8}} \hspace{0.6cm} (43.0, 46.3)  \\
Gemini 1.5 Pro & 40.1 \hspace{0.6cm} (38.3, 41.5)  \\
Claude 3.5 Sonnet & 43.1 \hspace{0.6cm} (41.4, 44.8)  \\
\hline
\end{tabular}
\caption{We report the decision-type weighted F1-score on analysis generation based on average precision and coverage@10. Appendix~\ref{appendix:f1_calc} has the calculation details.
}
\label{tab:results_main}
\end{table}

\xhdr{LMs have acceptable world knowledge}  
We find that LMs can identify some relevant conceptual variables based on the research question and dataset (i.e., a reasonable precision and coverage for conceptual variables). In \benchname, many relevant conceptual variables are  
possibly hinted at in the research question and available data columns. Although our setting is realistic and common, future work could explore how LM agents perform in generating hypotheses and identifying relevant data without such context~\cite{Majumder2024DatadrivenDW}.
In addition, we find that the best general LMs (i.e., Gemini-1.5 Pro, Mixtral-8x22b, Claude-3.5-Sonnet and GPT-4o) perform well on the MCQs (Fig.~\ref{fig:mcq}). They can discern the obvious transformations for a given conceptual variable. In contrast, code-specific LMs, like CodeLlama and DeepSeek-Coder, struggle to identify the correct decision. 

\figResultsType

\figResultsHumanEvalComparison

\xhdr{Most LMs can generate non-empty executable analyses}
For generating an analysis, we find that most large LMs can generate a non-empty executable analysis over 60\% of the time, with GPT-4o being the best at 96\% (Fig.~\ref{fig:result_type}). Among the open-source models, Mixtral-8x22b performs best, generating an executable analysis 73\% of the time and DeepSeek-Coder also does surprisingly well at 65\%. In a manual inspection of non-executable analyses, we notice issues with respect to hallucinating data attributes. Taking one of DeepSeek-Coder's submissions to the soccer dataset as an example, we observe plausible looking code, but it hallucinates the ``RefCountry'' column, which does not actually appear in the data table (Figure \ref{fig:ex-code-2}-7).

\xhdr{LMs struggle to specify statistical models and concretely operationalize conceptual variables}
LMs perform relatively poorly in forming statistical models with the right conceptual variables (precision below 35\%) and operationalizing the variables (precision below 60\%).
In addition, LMs perform even worse in terms of coverage for forming statistical models with conceptual variables (coverage@10 below 13\% across) and operationalizing the variables (coverage@10 below 27\%). This indicates there is room for improvement not only in generating valid analyses, but also generating more complex and diverse analyses that might require additional reasoning beyond the basic steps.

\xhdr{LMs are limited to forming basic analyses}
Figure~\ref{fig:result_type} also shows that  
many LM's submissions contain empty transform code, especially for GPT-3.5 Turbo and Gemini 1.5 Pro. We also observe low coverage of the ground truth examples (Fig.~\ref{fig:result_main} bottom), especially with respect to data transformations and specific model specifications. 
%\sandy{Re above, can you state why you think this happened?}
Through qualitatively reviewing a random sample of LM-generated analyses, we find that LMs  often perform basic analysis that can yield
%\sandy{Avoid imprecise words like "good." Try to be more specific or measurable.}
decent precision (i.e., matching basic decisions) but poor coverage across runs. See Appendix~\ref{appendix:qualitative} for examples. 

\xhdr{Agents can improve the diversity of analyses}
Comparing the one-turn and agent settings, LMs consistently had higher coverage when allowed to iteratively explore data. Moreover, 
%\sandy{Re below, why do you think this occurred?}
ReAct agents perform best overall on coverage for data transformations and statistical modeling, which require a more detailed understanding of data semantics (Fig.~\ref{fig:result_main} bottom). Future work can explore how augmenting agents with external knowledge (e.g., from academic papers) can further improve  performance.

\xhdr{Stronger performance on code generation does not translate directly to \benchname}
When comparing our results in analysis generation with those from the HumanEval coding benchmark (Fig.~\ref{fig:result_compare_humaneval}), we found that most metrics showed a positive correlation, indicating that higher HumanEval performance is broadly correlated with higher BLADE performance. However, coverage measures (Fig.~\ref{fig:result_compare_humaneval} bottom) had a weaker correlation compared to precision (Fig.~\ref{fig:result_compare_humaneval} top). This suggests that while current training methods, such as Reinforcement Learning from Human Feedback (RLHF) and instruction tuning, optimize for one solution, they may struggle to generate diverse solutions, a phenomenon observed in other contexts~\cite{Li2024PredictingVA}.

We also highlight that Gemini 1.5 Pro consistently performed better on precision than its HumanEval performance would suggest, while Mixtral-8x22B excelled in both precision and coverage for conceptual variables and data transformations. In contrast, CodeLlama consistently performed worse on BLADE than HumanEval. Given that Gemini 1.5 Pro and Mixtral-8x22B are general-purpose Mixture-of-Experts models, our findings highlight BLADE as a challenging benchmark that assesses more than just code generation. Our results identify specific areas for improvement, such as enhancing the complexity and diversity of analyses or generating justifiable statistical models.

\section{Related Work}
\label{sec:related-work}
Our work broadly relates to agent benchmarks for data science and LM agents in science.

\xhdr{Benchmarks for Data Science}
\label{sec:related-work-benchmark}
Many benchmarks~\cite{Li2024TapilotCrossingBA, Yin2022NaturalLT, Hu2024InfiAgentDABenchEA} assess agents’ data science code execution but are limited in measuring complex reasoning and external knowledge integration needed for scientific analyses. Other works focus on improving specific metrics in machine learning tasks~\cite{Huang2023MLAgentBenchEL, Hong2024DataIA} without evaluating intermediate decisions (examples in Table~\ref{tab:benchmark-comparison}). Some works assess agents’ causal~\cite{Jin2023CLadderAC} and quantitative reasoning~\cite{Liu2024AreLC}, but often lack data or involve closed-ended solutions, missing the flexibility inherent in open-ended scientific analyses. DiscoveryBench~\cite{Majumder2024DiscoveryBenchTD} evaluates agents on generating data-driven hypotheses but does not explicitly measure decisions in the analyses. In contrast, \benchname~focuses on evaluating \textit{multiple} valid analysis \textit{approaches} to \textit{open-ended}, \textit{data-driven} research questions.

\xhdr{Agents for Science}
\label{sec:related-work-agents}
Advancements in LMs have ignited research interest in applying agents to automate scientific discovery~\cite{Liang2023CanLL, RomeraParedes2023MathematicalDF, Shojaee2024LLMSRSE, Kramer2023AutomatedSD, Bran2023AugmentingLL, Boiko2023AutonomousCR, Majumder2024DatadrivenDW}. 
Our work seeks to provide a thorough automated evaluation of agents for scientific analyses across domains. 

\section{Conclusion}
We introduce \benchname, a benchmark designed to advance the development of LM agents for data-driven scientific tasks. We collected a dataset of research questions and data tables, along with ground truth analyses from expert annotators. To support an \textit{automatic}, \textit{decision-based}, and \textit{input-flexible} evaluation, we devised representations of core analysis decisions and developed corresponding matching algorithms. Although current generations of LMs can generate \textit{some} analyses matching the ground truth \textit{sometimes}, we find that these analyses are limited in complexity and lack diversity. 

\section{Limitations}
Our work is not without limitations. First, \benchname~does not evaluate an agent's ability to interpret the results of data analyses as part of the end-to-end data analysis process. Understanding and interpreting model results is vital but can be difficult to capture cleanly since it may require analysts’ subjective interpretation of the problem with respect to model results. We leave this important dimension for future work. 

In addition, though our work elucidates the decisions an agent may make, we do not explicitly evaluate the exploratory parts of an analysis. Further, we assume that the dataset is contained in a single, potentially extremely large table. This may not be common of all research datasets, but we believe this factor does not significantly reduce the scope of \benchname~since joining tables to enable downstream analyses is a task that LMs already commonly perform~\cite{Liu2023ACE, Li2024CodeSTB, Pourreza2023DINSQLDI}. 

Finally, some components of our evaluation rely on LMs (e.g., conversion of code to discrete transforms, semantically matching model, and conceptual variable), which are known to hallucinate. Therefore, we made multiple efforts to validate each component and do not think that hallucination  
significantly impacts our ability to effectively and automatically evaluate agents. We also open source these evaluation modules so that researchers can build upon them to improve our evaluation.

\section*{Acknowledgments}

We are grateful for the expert annotators who worked with us over multiple rounds to gather the highest quality data. We also thank the UW Behavioral Data Science Group members for their suggestions and feedback. Additionally, we thank Tiffany Zheng for her support and brainstorming on the figures, Josh Gardner and Andrew McNutt for their early feedback on the overall project, and Farhan Samir for his input towards evaluation. This research was supported in part by NSF grant IIS-1901386, NSF CAREER IIS-2142794, Bill \& Melinda Gates Foundation (INV-004841), and the Office of Naval Research (\#N00014-21-1-2154).

 % You can use another style if you prefer
\newpage
\bibliography{custom}
\clearpage
\appendix
\section{Appendix}
\label{sec:appendix}
\subsection{Data Collection Recruitment}
\label{appendix:recruitment}
Data analysts were recruited through open calls on social media platforms and personal connections. Of the analysts interested, a subset was selected based on their CVs reflecting education, training, and practices in statistical foundations and data analysis. The selected analysts provided sufficiently detailed analysis reports in a screening task and proceeded to the formal annotation phase. A total of 11 analysts participated in the final annotation (see Table \ref{tab:annotator_background} for annotator information).

The participating analysts self-reported an average of 6 years of experience in data analysis (range: 4-8 years), with 4 analysts performing data analysis on a daily basis and 7 engaging in it a few times a week. The team included 6 people holding or pursuing Ph.D. degree in Statistics or a related field (Ph. D. in Biostatistics, Ph.D. in Biomedical and Health Informatics, Ph.D. in Measurement, Evaluation, \& Research Methodology), the rest held at least 1 Master's degree in a related field. The analysts' occupations varied, 7 were graduate students, 3 held data scientist positions in the finance and technology industries, and 1 was a quantitative researcher in finance.

By assembling a team of analysts with diverse backgrounds and a broad range of expertise in statistical analysis methods, we ensure that the ground truth dataset is constructed using a comprehensive set of methods. At least half (n=5) of the analysts self-reported being familiar ``to a high extent" or ``to a very high extent" with common classes of analysis methods including descriptive statistics, inferential statistics, hypothesis testing, estimation, correlation, and regression.

\subsection{Data Collection Procedure}
\label{appendix:annotation_training}
While the analysts were free to conduct the analysis in their preferred computational environment, we took several additional steps to ensure the quality of our ground truth.

To ensure consistency of annotations, we built a pipeline with structured training and annotation procedure aimed at ensuring well-prepared analysts, consistent and reliable analysis decision specifications, and a diverse range of justifiable models and analysis approaches. These decisions cover conceptual variable formulation, executing data transformations to operationalize the variables, and implementing statistical models (Sec.~\ref{sec:benchmark_tasks}).

To streamline the annotation process and reduce some of the cognitive load in specification, we developed a customized annotation interface that supports structured inputs and sanity checks.

We started with a training and familiarization procedure for the analysts. The process involved on-boarding and training to establish a clear mutual understanding of the expected level of analysis and the format of decision inputs to be recorded in the ground truth. We provided analysts with video and text tutorials, accompanied by a toy example implemented within the system. Multiple ad hoc meetings and Q\&A sessions were also held to further clarify the process and address any issues. Analysts were introduced to example crowd-sourced analyses~\cite{Schweinsberg2021SameDD, Silberzahn2018ManyAO} to align their mental models with justifiable alternative decisions and the model quality level.

Collaborative efforts were encouraged in curating and shaping the datasets, research questions, and meta-information. In the review and revision phase, we shared input from other annotators and presented LM-generated examples ($n\approx40$ per annotator,  per dataset) for analysts to label as correct or incorrect. This process helped identify gaps, promote diversity, and encourage the incorporation of additional justifiable decisions. Analysts labeled the generated examples as justifiable or not justifiable, drawing inspiration from their peers and LM-generated outputs. The diversity in familiarity with various analysis methods among the analysts complemented each other, resulting in a more robust set of annotations.

At the end of the annotation, we collected 118 conceptual variable decisions, 246 discrete transform decisions, and 172 modeling decisions (i.e., choice of statistical model and model formula).

\subsection{Analysis Decision Representations}
\label{appendix:matching_def}
In this section, we formally describe the representation of different analysis decisions as described in Section~\ref{sec:enable_eval}. These representations capture all alternative approaches in our ground truth and are matched with an agent's generated analysis artifacts (Sec.~\ref{sec:task_gen}).

\xhdr{Data Transformations}  Formally, a transform data flow graph is a bipartite graph with two types of nodes: \textit{transform} nodes and \textit{column pointer} nodes.
\begin{equation} G = (T \cup P,~E) \end{equation}
where~$T= \{t_1, t_2, \ldots\}$ is the set of transforms. $P= \{p_1, p_2, \ldots\}$ is the set of column pointers, and the set of edges is denoted by: \begin{equation} E \subseteq (T \times P) \cup (P \times T) \end{equation} 

Each transform $t$ represents one unit of transformation and is defined by a fixed set of transform verbs $V$ (Table \ref{tab:transform_verbs}). This set of transform verbs is based on existing data wrangling libraries \cite{arquero, satyanarayan2016vega}, were validated to cover every analysis decision in our benchmark, and represent a discrete data transformation decision that was made in wrangling the data.

Given our graph, we ultimately want to match based on the column values as a result of any series of transformations. Thus, each column pointer holds the column values and facilitates the flow of column values from transform to transform. We denote the column vector value at a column pointer node~$p$ as $\mathbf{v}_p$ and $S$ as the set of all column values associated with $G$.

\begin{equation} S= \{\mathbf{v}_p \mid p \in P\}\end{equation}

The set of input column pointers to a transform $t$ and the output column pointers from a transform $t$ are defined by $I(t)$ and $O(t)$: 
\begin{equation} 
I(t) = \{p \in P \mid (p, t) \in E \},
\end{equation}
\begin{equation} O(t) = \{p \in P \mid (t, p) \in E \}.\end{equation}

The exact transform performed dictates $I(t)$ and $O(t)$. Specifically, $O(t)$ reflects only the columns that are changed by $t$ and $I(t)$ are the columns that are necessary to compute the output $O(t)$ .

Our transform data flow graph satisfies the following properties:
\begin{align*}
|~I(t)| &> 0 \\ 
|O(t)| &\geq 1 \\
% \text{and} (|O(t)| = 1 or |O(t)| = |I(t)|) 
|I(p)| &= 1~\text{except for original columns}
\end{align*}

So far, a single data flow graph $G$, represents a unique series of transformations. To account for all alternative transformation choices (e.g., an alternative in which the filter step \circled{3} in Fig.~\ref{fig:transform_example} is skipped), we define $\eqAllGraphs =\{G_1, G_2, \ldots, G_n\}$ to be the set representing all unique series of transformations for an analysis. Note that any two graphs $G_i=(T_i, E_i)$ and $G_j=(T_j, E_j)$ may contain the same transformation (e.g., two graphs can contain the same derive rater average transform \circled{1}) and so $T_i \cap T_j \neq \emptyset$.

Finally, to keep track of all transformations across all justifiable alternatives, we define $\eqAllTransforms$  and $\eqColValues$ to be the set of all transformations and columns values, respectively, across all data flow graphs. Any agent benchmark submission will be matched against these ground-truth representations (described in Appendix~\ref{appendix:match_transforms}).

\begin{equation}
\boldsymbol{\mathcal{T}} = \bigcup_{i=1}^n T_i~~~~~~~~\eqColValues=\bigcup_{i=1}^n S_i
\end{equation}

\xhdr{Conceptual Variables} A conceptual variable $c \in \eqCvars$ is a triplet $(\eqCvarDesc, \eqCvarType, \eqCvarCols)$ where $\eqCvarDesc$ is a natural language description of the conceptual variable, $\eqCvarType \in \{\text{\texttt{IV}}, \text{\texttt{DV}}, \text{\texttt{Control}}\}$ is the variable type, and $\eqCvarCols \subseteq \eqColValues$ is the set of column vectors that operationalize $c$. Here, $\eqCvars$ denotes the set of conceptual variables across all alternative approaches.

\xhdr{Statistical Models} A statistical model $m \in \boldsymbol{M}$ is a tuple $(\eqModelDesc, \eqModelCols)$ where $\eqModelDesc$ is the natural language description of the statistical model and $\eqModelCols \subseteq \eqCvarCols$ is a set of column vectors associated with the model which are also associated with a conceptual variable. In addition, $\eqModelCols$ should be associated with only one series of transformations or one data flow graph, that is:
\begin{equation}
\exists S_i \in \{S_1, S_2, \ldots, S_n\} \mid \eqModelCols \subseteq S_i
\end{equation}
From $\eqModelCols$, we can also derive the associated conceptual variables $C_m \subseteq \eqCvars$ in a model. 

\begin{equation}
C_m = \{c_i \mid \eqCvarColsi \cap \eqModelCols \neq \emptyset\}
\end{equation}
In addition, for each statistical model $m$, there is one associated variable that is a DV, at least one associated variable that is an IV and 0 or more Control variables.
% DV $|\{ c_i \in C_m \mid r_i = \text{DV}\}| = 1$, at least one IV $|\{ c_i \in C_m \mid r_i = \text{IV}\}| \geq 1$, and 0 control variables.
$\boldsymbol{M}$ denotes the set of statistical models across all alternative approaches.

\figExampleSubmission

\subsection{Decision Matching Procedure}
\label{appendix:matching}
With an understanding of the representations of different analysis decisions, we now describe a procedure to match an agent-generated analysis to the ground truth.

Given the agent submission artifacts (Fig.~\ref{fig:example_submission}), we first apply LMs to handle the conversion of generated artifacts into our ground truth representation format. Specifically, we use GPT-4 to perform two tasks: convert the transform function into individual transform units, and translate the modeling function into a statistical model specification (e.g., linear regression) along with the columns used in the model (Fig.~\ref{fig:convert}). 

% \ken{this could also be finicky, how cane we explain that this is fine, validated and justfied. Before, tim mentioned this module is available to the public which people can improve on.} \tim{could say sth like: We manually evaluated this procedure by inspecting ten (or more?) generated representations, and found the performance to be sufficient for our purposes (10/10 correct). Full prompt is given in Appendix XXX.}
% \mike{I think this might be one of the biggest risks in your paper. One strategy is supporting the strength of this intermediate step by showing that your final results make sense. If there aren't major surprises in your conclusions that could be the result of bias in this steps then you'll have less to worry about}

Next, we describe the procedure to match a given analysis to the ground truth. Specifically, given the ground truth $\eqAllGraphs$, $\eqAllTransforms$, $\eqCvars$, and $\boldsymbol{M}$, we describe matching a single analysis containing $G'$, $T'$, $\eqCvars'$, and $\boldsymbol{M}'$.

\figConvert

\subsubsection{Matching Transforms}
\label{appendix:match_transforms}
Data transformations are inherently open-ended with multiple valid approaches and free-form responses. Our goal is to capture how well agents perform in the underlying data analysis decisions. Therefore, we define multiple approaches to capture different levels of performance (i.e., getting the exact column vector vs. rough same steps) in how well a given analysis matches with the decisions in the ground truth: value matching and graph matching.

In both matching schemes, we determine whether a match occurs (i.e., based on matching column values or the graph structure based on the transform specification) and match all upstream transformations based on the data flow graph $G$. 

Here, in order to evaluate the quality of a series of transform $T'$ in $G'$, we attempt to identify ground truth transforms $t \in \eqAllTransforms$ associated with  $\eqAllGraphs$ that matches with $t'\in T'$, that is, $Match(t)=1$ and $Match(t')=1$. 

To match all transforms $\eqAllTransforms$ in all specified alternatives with $T'$, as the transforms are situated in the graphs, we perform all pairwise matching between $G \in \mathcal{G}$ and $G'$.

\xhdr{Value Matching}
In value matching, we want to match two \textit{series} of transformations if they result in the same column value.

Given $S$ and $S'$ denoting the sets of column vectors associated with $G$ and $G'$, if $\mathbf{v}_p \in S = \mathbf{v}_{p'} \in S'$ (i.e., all cell values are equal when comparing two column vectors at column pointer nodes $p$ and $p'$), then this means that the series of transformations that resulted in $\mathbf{v}_p$ and $\mathbf{v}_{p'}$ are equivalent. Therefore, all parents transforms of $p$ in $G$ and $p'$ in $G'$ should be matched.

Let $I(p)^+$ denote the set of transforms in the transitive closure of $p$ and its ancestors: $I(p) = \{t \in T \mid T\in I(p) \text{ or } T \in I(I(I(p))) \ldots\}$.
If $\mathbf{v}_p = \mathbf{v}_{p'}$, then $I(p) \subseteq T$ is matched and $I(p') \subseteq T'$ are matched.

\begin{align*}
Match(t) &= 1~~\forall t \in I(p)^+~\text{and}  \\
Match(t') &= 1~~\forall t \in I(p')^+
\end{align*}

While it may be the case that there are other column values involved in $O(t)$ which may differ, we at least know for sure that two series of transformations produced the same column value. In addition, because the definition of each $t$ is set to only include the affected columns, we find that the match of values in two pairs of columns is a sufficient criterion for equivalence.

\xhdr{Fuzzy Graph Isomorphism Matching} Value matching may be considered to be too strict, especially when small changes in the numerical parameters of a transform can lead to different column values (e.g., in Fig.~\ref{fig:transform_example}, filter on \texttt{rpg > 0.5} vs. \texttt{rpg > 0.45}). To allow greater flexibility in the matching, we introduce fuzzy graph matching. In graph matching, we match based on the transform verbs and column specifications rather than the exact column values (e.g., choosing to filter on \texttt{rpg} and \texttt{r\_avg} after steps \circled{1} and \circled{2}). More specifically, if two series transforms shared the same high-level definition in which transforms are used in a similar way defined by the transform verb and parameter columns and dataflow, then they should be equivalent. 

To accomplish this, we add a node label mapping $L: T \rightarrow V \times P^n$ mapping the transform to its associated transform verb and column pointer parameters (e.g., step \circled{3} in Fig.~\ref{fig:transform_example} would have the node label $(\text{filter}, \{p_{\texttt{rpg}}, p_{\texttt{r\_avg}}\})$ where $p_\texttt{rpg}$ is the column node associated with the \texttt{rpg} column and $p_{\texttt{r\_avg}}$ is the column node associated with the \texttt{r\_avg} column). Given this definition, if a subgraph is equivalent to another subgraph, then this means they represent the same choices of transforms (at a higher-level of abstraction relative to Value Matching). 

More formally, let $H(t)$ denote the subgraph induced by the transitive closure of $t$ and its parents. $H(t)$ captures both the transform nodes and the relevant column pointer nodes.
% Specifically, let $p \in O(t)$, then the nodes of the graph $H(t)$ are denoted by $I(p)^+ \cup \{p \in P \mid (p, t) \in E \text{ or } (t^, p) \in E \text{ and } t, t^ \in I(p)^+\}$. The edges are those induced by these nodes.
If $H(t)$ is isomorphic to $H(t')$, including the node labels added from $L$ and $L'$, then all $t$ in $H(t)$ and $t$ in $H(t')$ are matched.  

\begin{align*}
Match(t) &= 1~~\forall t~~\text{in the graph}~~H(t) \\
Match(t') &= 1~~\forall t~~\text{in the graph}~~H(t') \\
\end{align*}

\subsubsection{Matching Conceptual Variables}
\label{appendix:cv_match}
Given $c \in \eqCvars$ and $c' \in \eqCvars'$, $c$ and $c'$ are equivalent if $\eqCvarType = \eqCvarType'$ and $\eqCvarDesc$ and $\eqCvarDesc'$ are semantically equivalent. For practical purposes, we use a language model to determine semantic equivalence. Specifically, we use GPT-4o following~\citeposs{Liang2023CanLL} prompting approach. 

We input JSON-formatted conceptual variable specifications for $\{\eqCvarDesc ~|~c \in \eqCvars\}$ and $\{\eqCvarDesc ~|~c \in \eqCvars'\}$. The LM then generates a JSON output where containing the pair of matching point IDs, and an associated similarity value providing the explanation for the match (see Fig.~\ref{fig:prompt5} for the prompt). 

% \ken{what sentence to say this is okay to do and does good enough?} \tim{same as previously. say you manually evaluated N examples and found this to work very well X/Y correct. or similar}
% \mike{How much can you rely on Liang et. al? If you use something very close to their method, which they validate, then say that}
% \ken{give a name for this LM matcher? "conceptual variable matcher"?}

% and validate this with expert human matching evaluation, achieving an accuracy of $X$.

\subsubsection{Matching Statistical Models} 
\label{appendix:model_match}

Given $m \in \boldsymbol{M}$ and $m' \in \boldsymbol{M}'$, we define two levels of matching: semantic and conceptual model-based matching. First, $m$ and $m'$ are semantically matched if $\eqModelDesc$ and $\eqModelDesc'$ are semantically equivalent following the same matching procedure for conceptual variables (see Fig.~\ref{fig:prompt6} for the prompt). This represents a coarse level of matching.

As determining the choice of a justifiable model involves including the right conceptual variables in the model, we then perform matching based on the conceptual variables (Appendix.~\ref{appendix:cv_match}) associated with the model.

\subsection{Baseline ReAct Agent Details}
\label{appendix:ReAct}

The baseline framework is an ReAct agent with~\texttt{[Thought]}, \texttt{[Action]}, and ~\texttt{[Observation]} stages before a final \texttt{[Finish]} stage. The initial \texttt{[Thought]} stage integrates the current context (i.e., latest observation) and the prior outputs (i.e., history of thoughts, actions, and observations) to formulate the next step action. Next, with the ~\texttt{[Action]} tag, the LM calls the underlying notebook and executes a new cell with the new LM-generated code. The \texttt{[Observation]} then comes from the notebook environment and is the string representation of the last-line output in the code following~\citealp{Yin2022NaturalLT}. This cycle repeats until the LM decides to output the final analysis with the \texttt{[Finish]} tag. The prompt for the agent includes one example of a ReAct trajectory (\texttt{[Thought]} ->~\texttt{[Action]} -> ~\texttt{[Observation]}) that iteratively explores the data. See Figure~\ref{fig:prompt9} for the prompt template.

The notebook sandbox environment uses Python 3.10 with the following imports:
\begin{lstlisting}[caption={}, label={lst:rcode}]
import pandas as pd
import sklearn
import scipy
import statsmodels.api as sm
import statsmodels.formula.api as smf
import numpy as np
import matplotlib.pyplot as plt
import seaborn as sns
\end{lstlisting}

These imports were determined during development such that the code generations do not involve any import errors on the main coding libraries.

Compared to the one-turn setting, the ReAct agent can explore the data more closely. In our experiments, we allow the agent to perform up to 10 steps, interacting with the environment with the full context of prior actions and observations.  Based on preliminary experiments, we determined that the ReAct agent needed LMs with at least an 8k context window to handle multiple turns of code execution outputs. Because of this, we performed experiments with the ReAct framework on the following LMs: Mixtral-8x22b, GPT-3.5 Turbo, GPT-4o, Gemini 1.5 Pro, and Claude 3.5 Sonnet.

\subsection{Prompt Templates}
\label{appendix:propmts}
The following figures (Figures \ref{fig:prompt1}-\ref{fig:prompt7}) show the various prompt templates used in the construction and evaluation of \benchname. The prompts in Figure~\ref{fig:prompt1} and Figure~\ref{fig:prompt2} are used to elicit alternative conceptual variables and data transformations in the benchmark data collection (Sec.~\ref{sec:benchmark_consturction}). 

The next set of prompts are used in the automatic evaluation of \benchname (Figures~\ref{fig:prompt5}-\ref{fig:prompt4}).
Figure~\ref{fig:prompt5} shows the prompt to semantically match conceptual variables. 
Figure~\ref{fig:prompt3_1} and~\ref{fig:prompt3_2} show the prompt for converting the agent's transformation function submission (e.g., Fig.~\ref{fig:example_submission}). Figure~\ref{fig:prompt5a} shows the prompt to convert the statistical modeling function into a natural language specification of the model and the columns in the transformed data table that are used in modeling. Finally, Figure~\ref{fig:prompt6} shows the prompt used to semantically match statistical models.

We also include the prompts for our evaluation tasks. Figure~\ref{fig:prompt4} shows the instructions to generate the entire analysis, while Figure~\ref{fig:prompt9} shows our implementation of the ReAct framework, which guides an AI assistant through reasoning and action steps for data analysis tasks. Figure~\ref{fig:prompt7} gives an example of our MCQ prompt. 

Most of these prompts utilize a JSON representation of Pydantic objects for standardized formatting, leveraging Langchain's Pydantic parser\footnote{\url{https://python.langchain.com/v0.1/docs/}}. Additionally, the schema of the dataset is represented as a JSON object, generated using the data summarizer from~\citealp{dibia2023lida}. Figure~\ref{fig:prompt3_2} provides a detailed description of the transformation API used in the prompt for Figure~\ref{fig:prompt3_1}, specifying the available transformation verbs and their corresponding input/output mappings. Figure~\ref{fig:prompt8} provides the one-shot example to guide the LM in generating an analysis (i.e,. the prompt in Fig.~\ref{fig:prompt4}).

\subsection{Evaluation Metrics Details}
\label{appendix:f1_calc}
% For a given LM agent, we take $m$ independent runs from which we calculate our metrics from. Here, we discuss the metrics in detail.

\xhdr{Average Precision}
Average precision is calculated as the mean of the precision scores across all individual runs. For a decision type (i.e., conceptual variables, transformations, statistical modeling) and a given set of agent-submitted decisions for runs \( \{R_1, R_2, \dots, R_n\} \) with a corresponding ground truth set $G$, the precision for each run $R_i$ is calculated as:

\begin{equation}
\text{Precision}(R_i) = \frac{|R_i \cap G|}{|R_i|}
\end{equation}

The average precision $p_{avg}$ is then computed as:
\begin{equation}
p_{\text{avg}} = \frac{1}{n} \sum_{i=1}^{n} \text{Precision}(R_i)
\end{equation}

\xhdr{Coverage@k}
Coverage@k is defined as the proportion of the ground truth set that is covered by the union of items across a sample of \textit{k} randomly selected runs (assuming the total number of runs $n > k$. Specifically, for a decision type and agent submitted decisions across a sample of $k$ runs \( \{R_1, R_2, \dots, R_{k}\} \), $coverage@k$ is calculated as:
\begin{equation}
{coverage}@{k} = \frac{\left|\bigcup_{i=1}^{k} R_i \cap G \right|}{|G|}
\end{equation}

For modeling decisions in which each run has one submission, the denominator is $\text{min}(|G|, k)$. 

In our experiments, we report coverage@10 for several reasons. First, we manually determined that for all datasets in~\benchname, conceptual variable and transformation decisions can be adequately covered in 10 runs. In addition, generating 10 independent analyses represents a reasonable and realistic scenario, mirroring a situation where one might leverage crowd-sourced analyses from 10 different analysts.

\xhdr{F1-score}
To reflect the overall performance while balancing precision and coverage, we compute F1-score calculated as follows:
\begin{equation}
F1 = \frac{2 \times (p_{\text{avg}} \times coverage@k)}{p_{\text{avg}} + coverage@k}
\end{equation}

To capture performance on~\benchname~in a single metric, for each decision type, we first take $p_\text{avg}$ and $coverage@10$ averaged across all datasets and calculate the F1-score. Next, we take the \textit{weighted-averaged} F1-score based on the number of ground truth decisions for each decision type. For statistical modeling decisions, the weight is based on $\text{min}(|G_{model}|, 10)$. 

\xhdr{Bootstrap Estimates and Confidence Intervals}
To account for the variability in selecting subsets of runs (especially for computing coverage@10), we employed a bootstrap procedure to estimate the expected F1-score and its confidence intervals. Specifically, we performed  $m=1000$  iterations of random sampling with replacement from the set of runs for each dataset. In each iteration, we recalculated both average precision and coverage@10, and then computed the corresponding F1-score. The final reported F1-score is the average of these bootstrap iterations, with a 95\% confidence interval derived from the distribution of the bootstrap samples.

% This approach ensures that our metric robustly captures the balance between precision and coverage, reflecting the system’s performance in a realistic and interpretable manner.

\subsection{Case Studies with Qualitative Insights}
\label{appendix:qualitative}
\begin{figure*}[p]
  \centering
  \includegraphics[width=\textwidth]{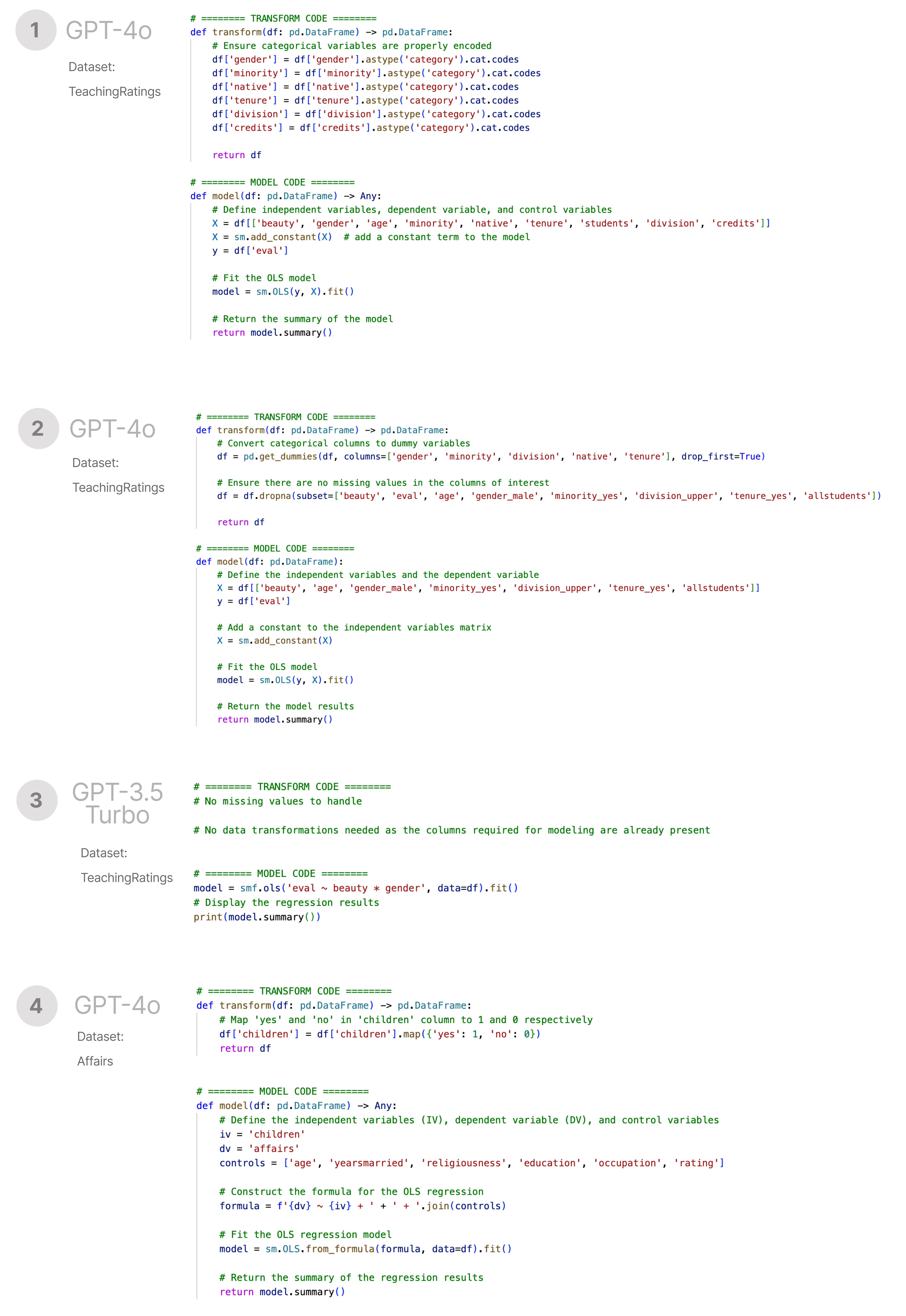}
  \caption{Part I for examples of LM-generated python codes transformations and models from case studies. Model types and corresponding datasets are shown on the left of the code.}
  \label{fig:ex-code-1}
\end{figure*}

\begin{figure*}[p]
  \centering
  \includegraphics[width=\textwidth]{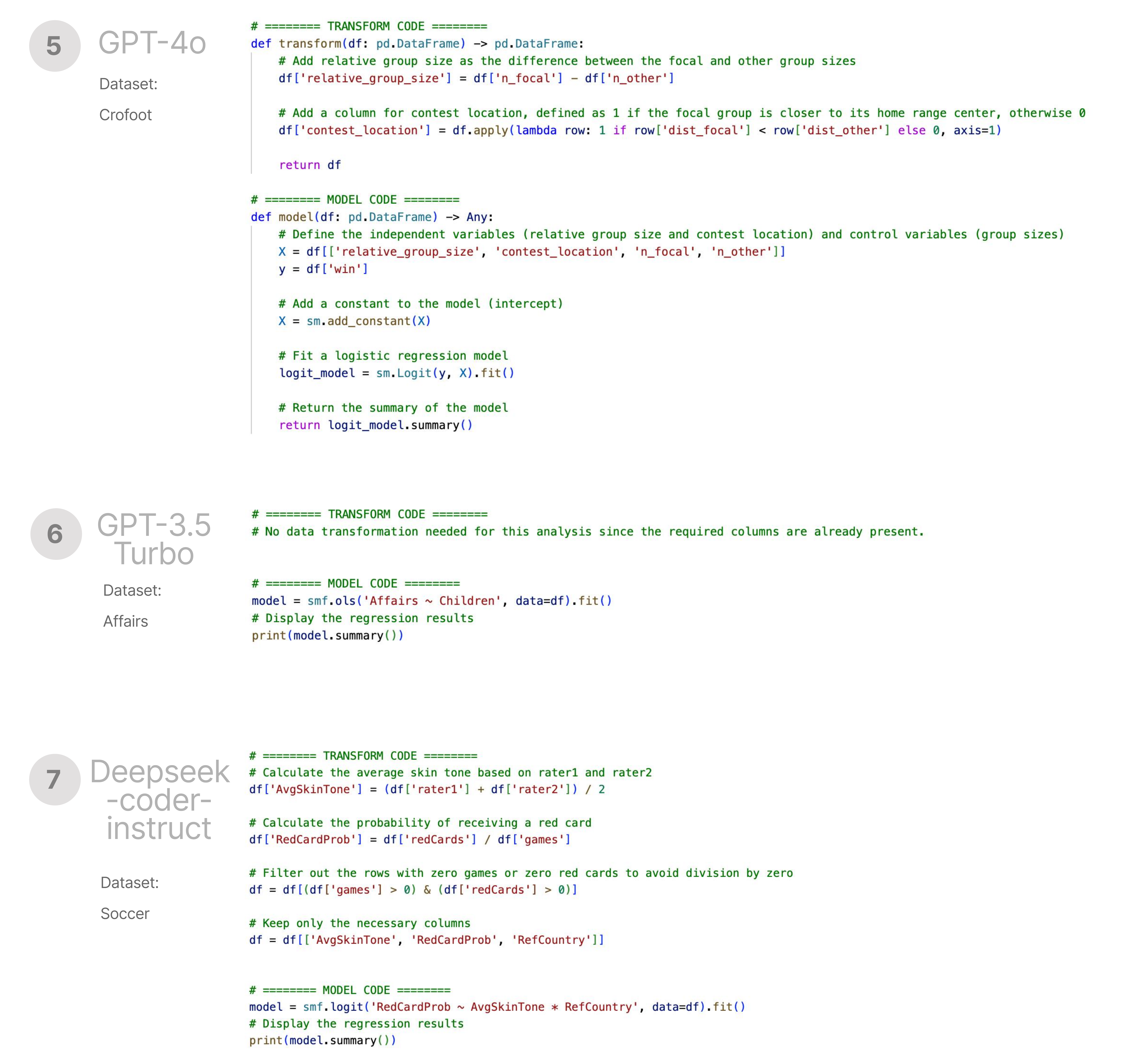}
  \caption{Part II for examples of LM-generated python codes transformations and models from case studies. Model types and corresponding datasets are shown on the left of the code.}
  \label{fig:ex-code-2}
\end{figure*}

To gain additional insight into the performance of LMs, two of the annotators sampled 56 output files from LM-generated results for qualitative case studies. Our findings reveal several limitations in LMs' ability to generate robust and reliable analyses:
\begin{enumerate}
\item \textbf{Composite Variables:} In the TeachingRatings dataset (Figure \ref{fig:ex-code-1}-1, Figure \ref{fig:ex-code-1}-2), GPT-4 failed to create important composite variables, such as evaluation response rate, despite their interpretability and explanatory power. LLMs often included only one of the component variables.
\item \textbf{Interaction Effects:} GPT-3.5 (Figure \ref{fig:ex-code-1}-3) struggled with understanding interaction effects in linear regression models, often including irrational interaction terms without main effects (e.g., \texttt{eval \textasciitilde{} beauty * gender}).
\item \textbf{Variable Selection:} While GPT-4 provided more comprehensive models with most control variables (see one example in Figure \ref{fig:ex-code-1}-4), it sometimes included redundant variables (e.g., ``relative group size'' derived from ``n\_focal'' and ``n\_other'') (Figure \ref{fig:ex-code-2}-5). In contrast, GPT-3.5 often used very minimal models (only one IV with no controls) (Figure \ref{fig:ex-code-2}-6). 
\end{enumerate}

% \newpage
\datasetmetadata

\tabHorizontalComparison
% \newpage
\annotatorinfo
\taxonomyverbs

\newpage
\begin{figure*}[p]
  \centering
  \includegraphics[width=0.9\textwidth]{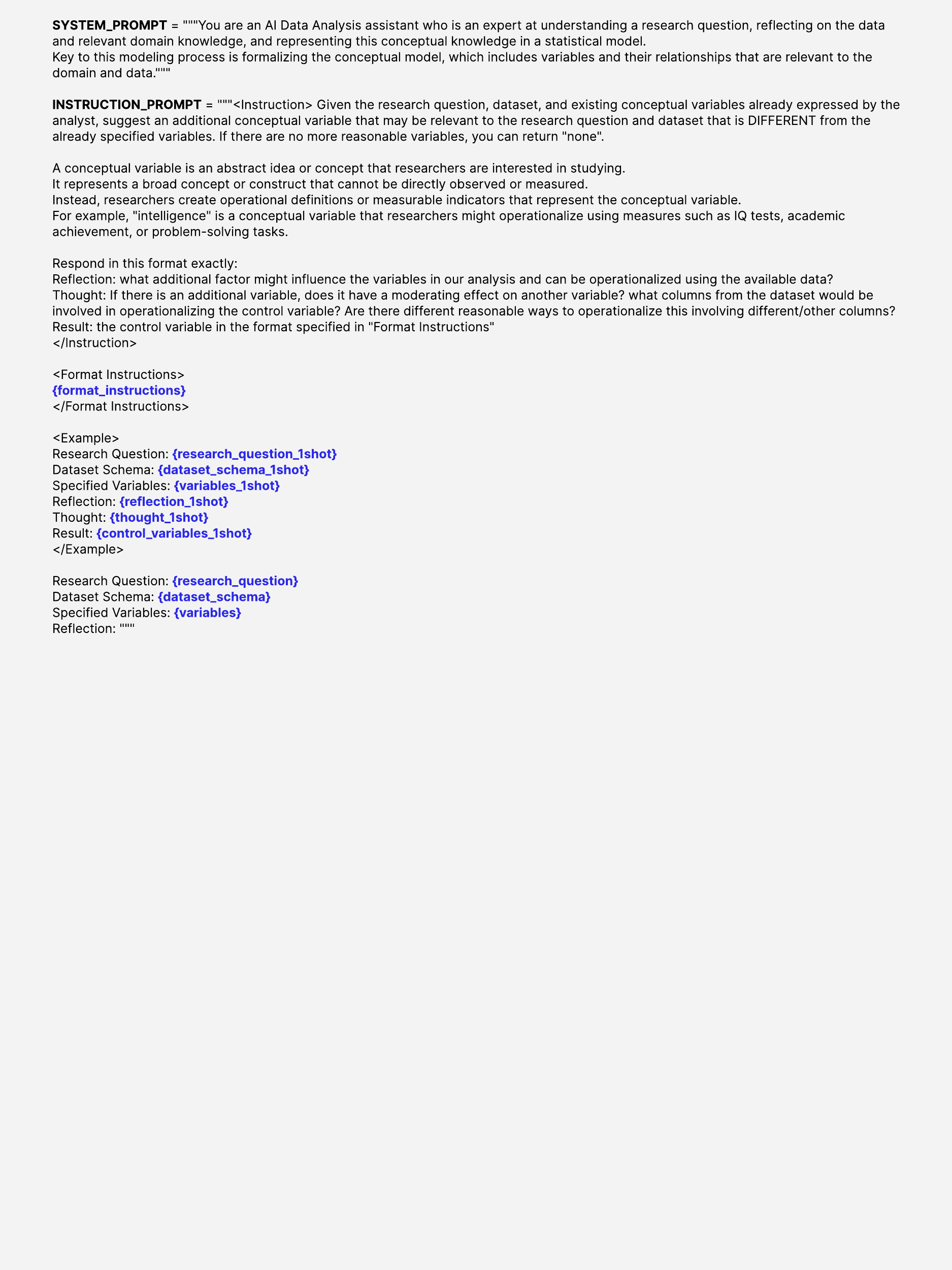}
  \caption{Prompt template A asking the LM to suggest an additional conceptual variable relevant to the research question and dataset. The format instructions asks the LM to generate a JSON representation of a Pydantic Object. Specifically we use Langchain’s pydantic parser (\url{https://python.langchain.com/v0.1/docs/modules/model_io/output_parsers/types/pydantic/}) for the format instructions. The dataset schema is a, JSON representation of a data table. We use the data summarizer in LIDA~\cite{dibia2023lida}}
  \label{fig:prompt1}
\end{figure*}

% \footnote{The format instructions asks the LM to generate a JSON representation of a Pydantic Object. Specifically we use Langchain’s pydantic parser (\url{https://python.langchain.com/v0.1/docs/modules/model_io/output_parsers/types/pydantic/}) for the format instructions. The dataset schema is a, JSON represetnation of a data table. We use the data summarizer in LIDA~\cite{dibia2023lida}}

\newpage
\begin{figure*}[p]
  \centering
  \includegraphics[width=0.9\textwidth]{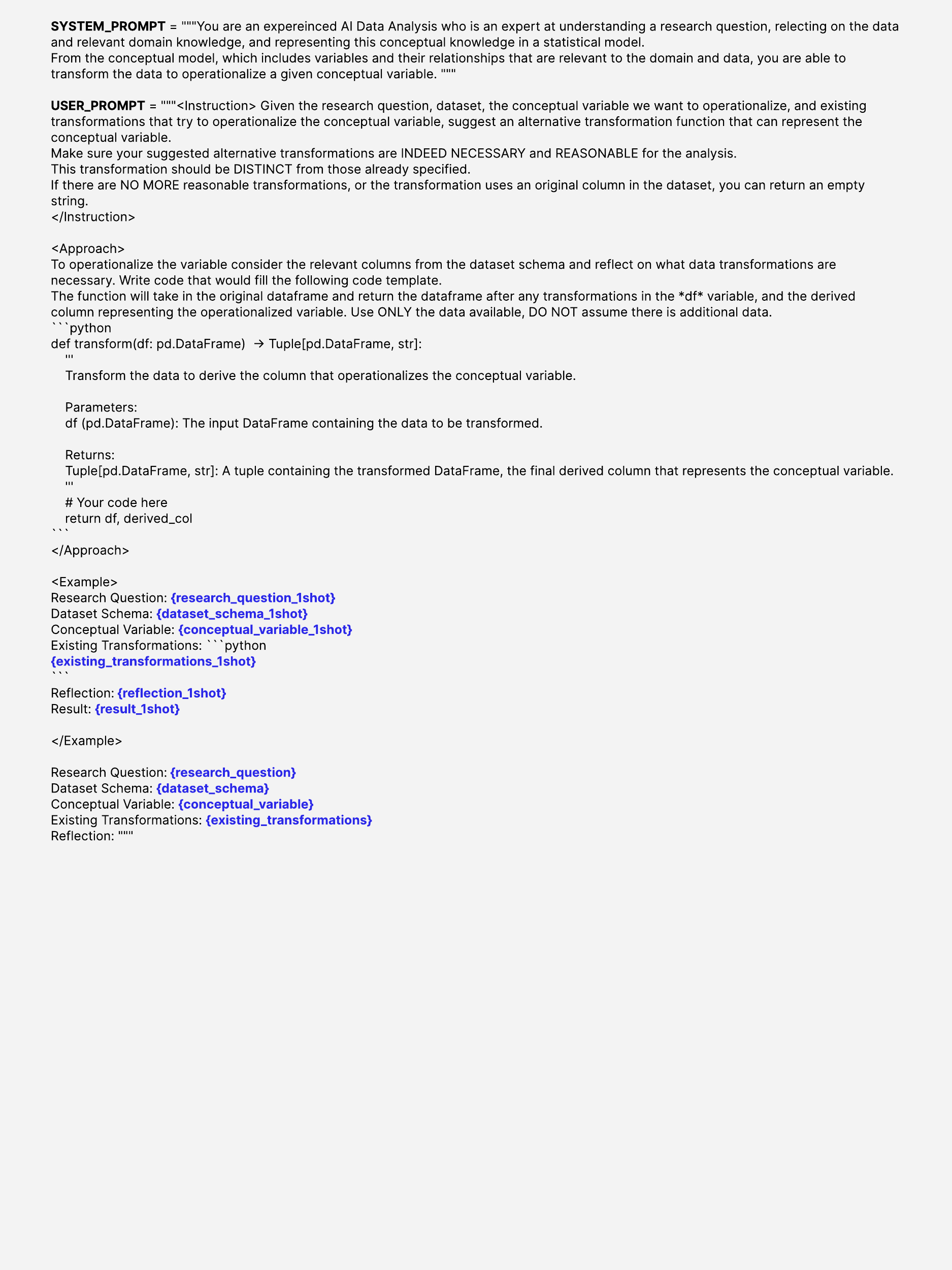}
  \caption{Prompt template B asking the LM to suggest an alternative transformation in Python that transforms the given data columns to operationalize a conceptual variable.}
  \label{fig:prompt2}
\end{figure*}

\newpage
\begin{figure*}[p]
  \centering
  \includegraphics[width=0.9\textwidth]{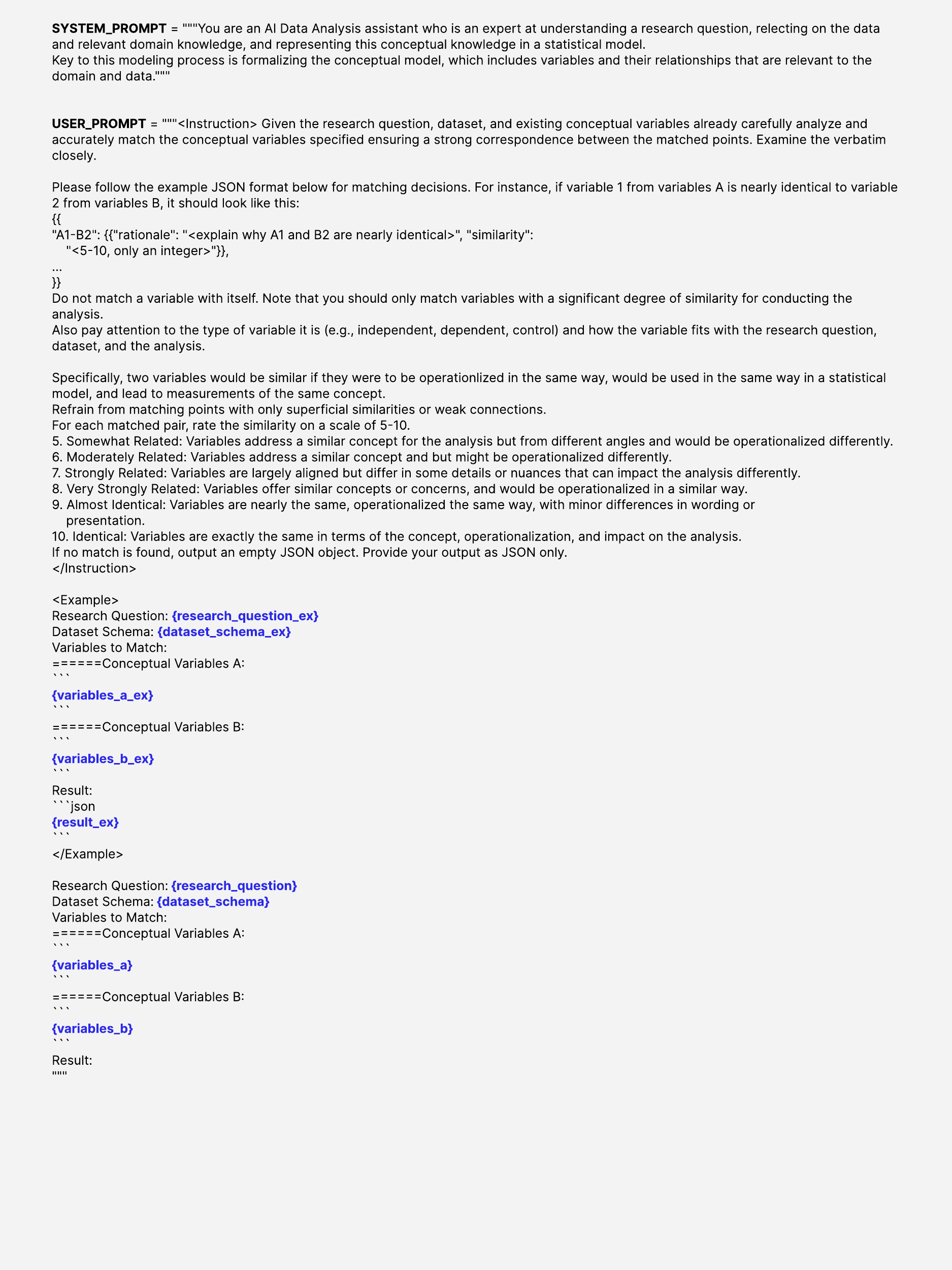}
  \caption{Prompt template C asking the LM to match conceptual variables from two given sets, considering their similarity in the context of the research question and dataset.}
  \label{fig:prompt5}
\end{figure*}

\newpage
\begin{figure*}[p]
  \centering
  \includegraphics[width=0.9\textwidth]{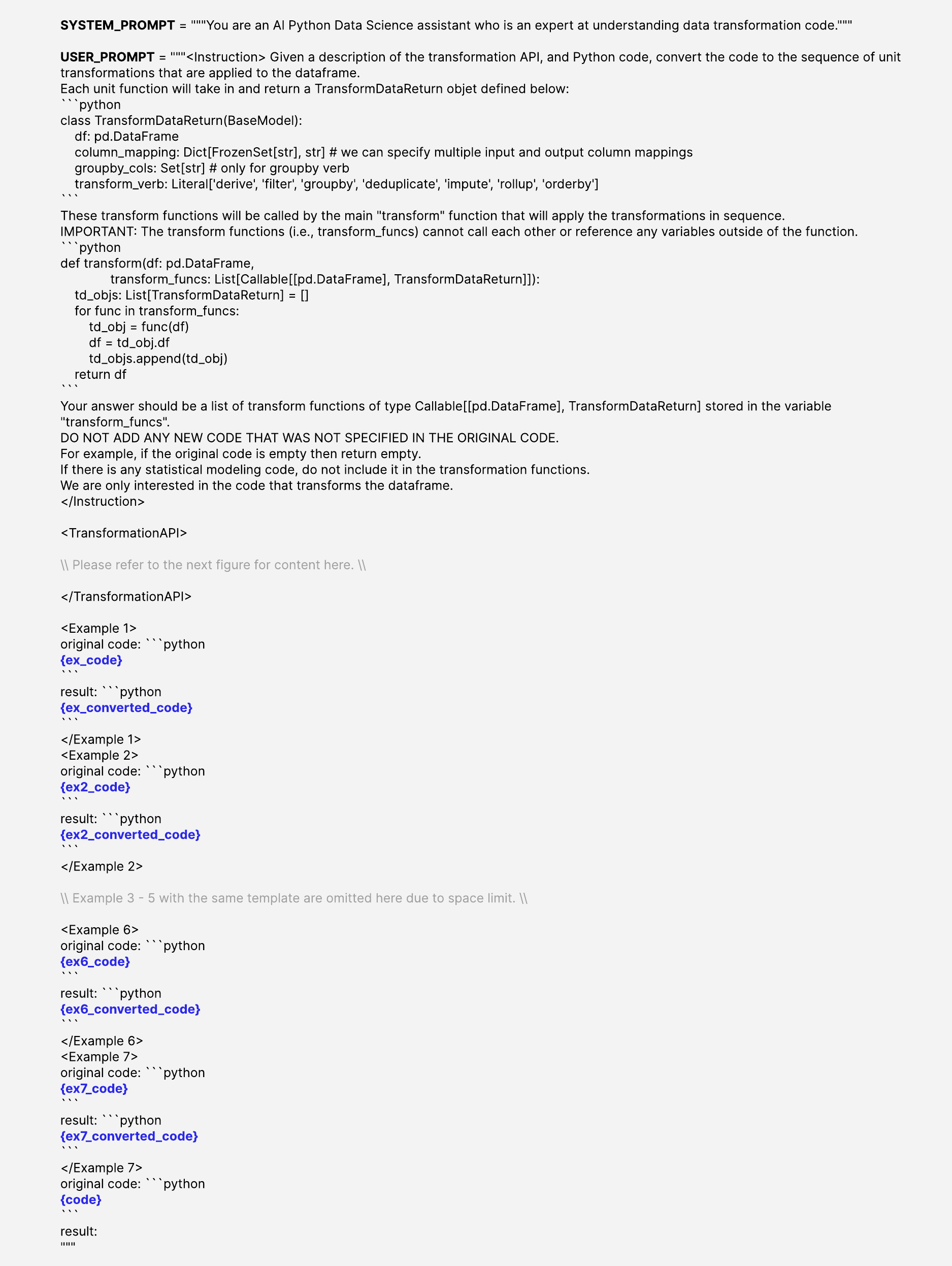}
  \caption{Prompt template D asking the LM to convert a given Python function for data transformation into a sequence of unit transformation functions, each taking a DataFrame as input and returning a TransformDataReturn object. Refer to Figure \ref{fig:prompt3_2} for the content of TransformationAPI.}
  \label{fig:prompt3_1}
\end{figure*}

\newpage
\begin{figure*}[p]
  \centering
  \includegraphics[width=0.9\textwidth]{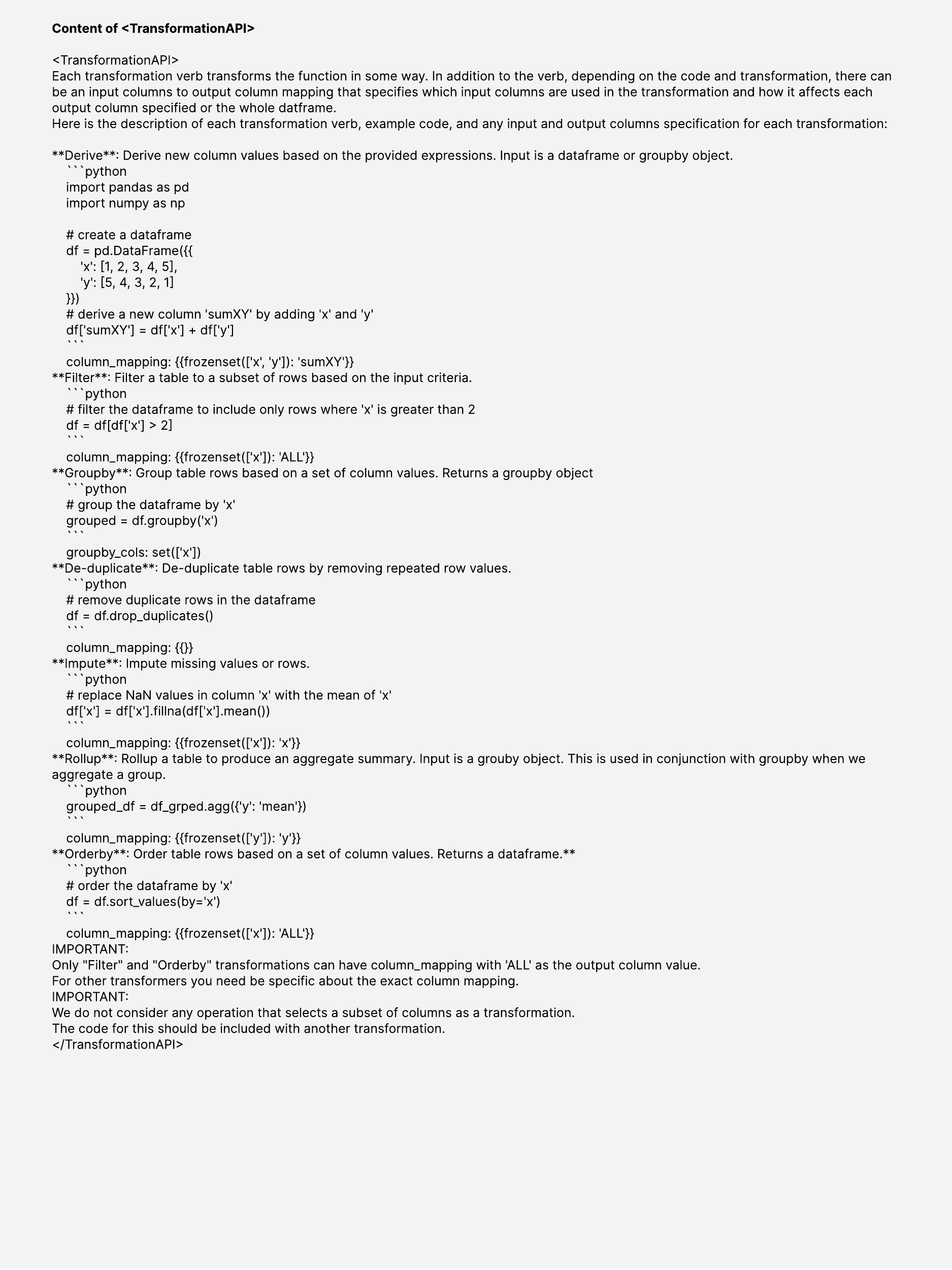}
  \caption{Detailed description of the transformation API, specifying the available transformation verbs (derive, filter, groupby, de-duplicate, impute, rollup, and orderby) along with example code and input/output column mappings for each transformation. This is used in part of the prompt in Figure \ref{fig:prompt3_1}.}
  \label{fig:prompt3_2}
\end{figure*}

\newpage
\begin{figure*}[p]
  \centering
  \includegraphics[width=0.9\textwidth]{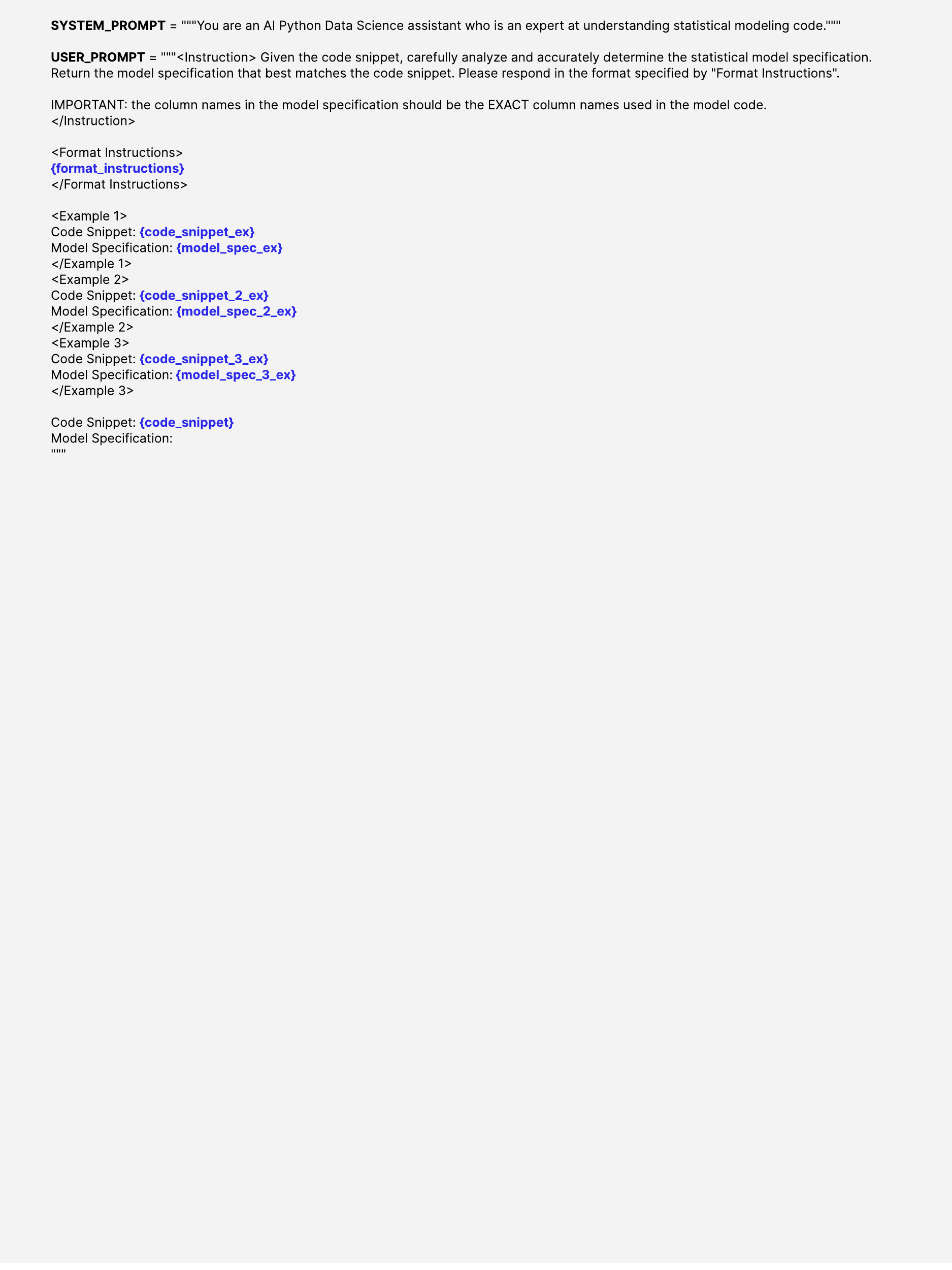}
  \caption{Prompt template E instructing the LM to analyze code snippets and determine the corresponding statistical model specifications.}
  \label{fig:prompt5a}
\end{figure*}

\newpage
\begin{figure*}[p]
  \centering
  \includegraphics[width=0.9\textwidth]{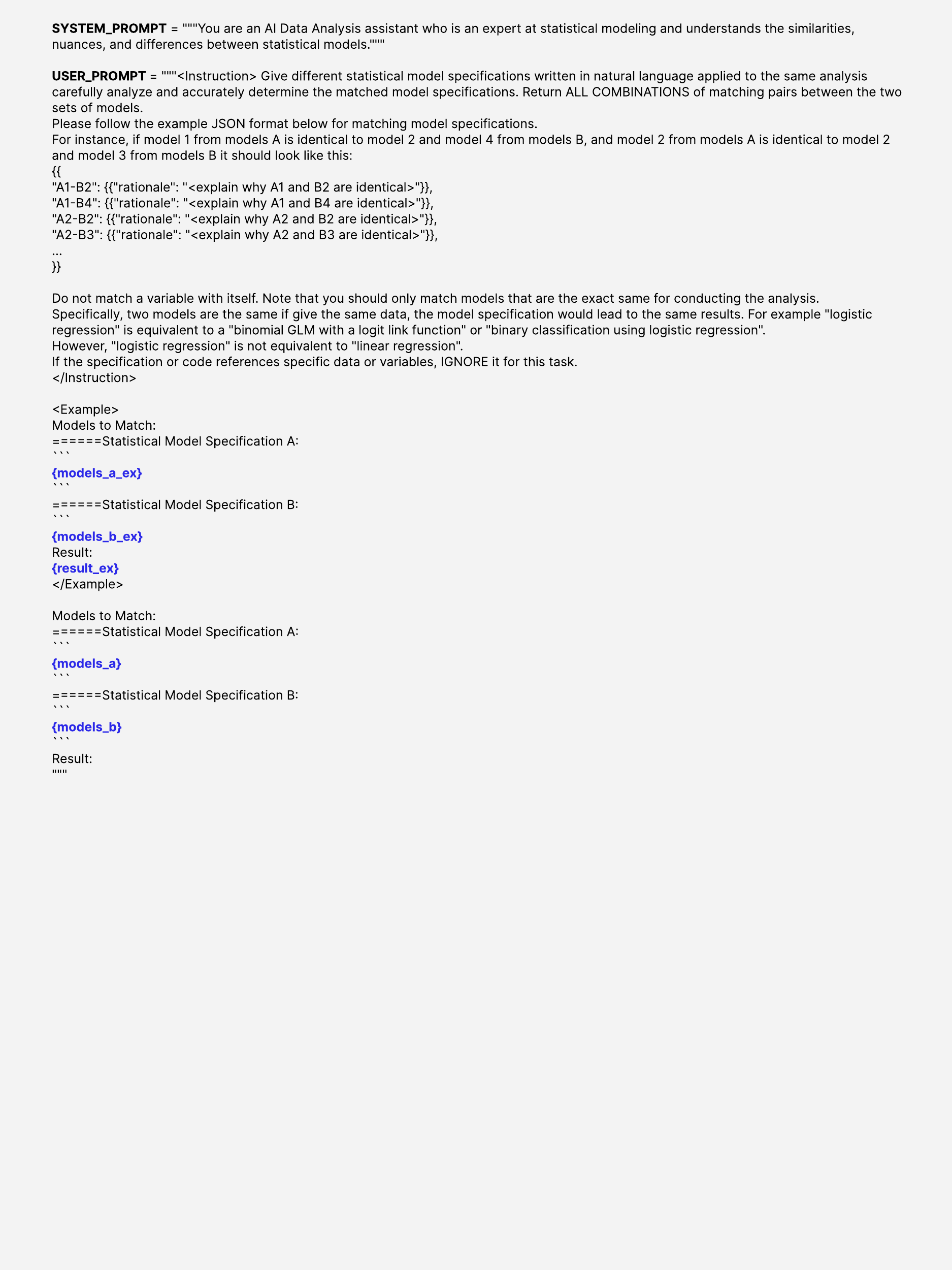}
  \caption{Prompt template F asking the LM to match statistical model specifications written in natural language, determining which models from two given sets are identical.}
  \label{fig:prompt6}
\end{figure*}

\newpage
\begin{figure*}[p]
  \centering
  \includegraphics[width=0.9\textwidth]{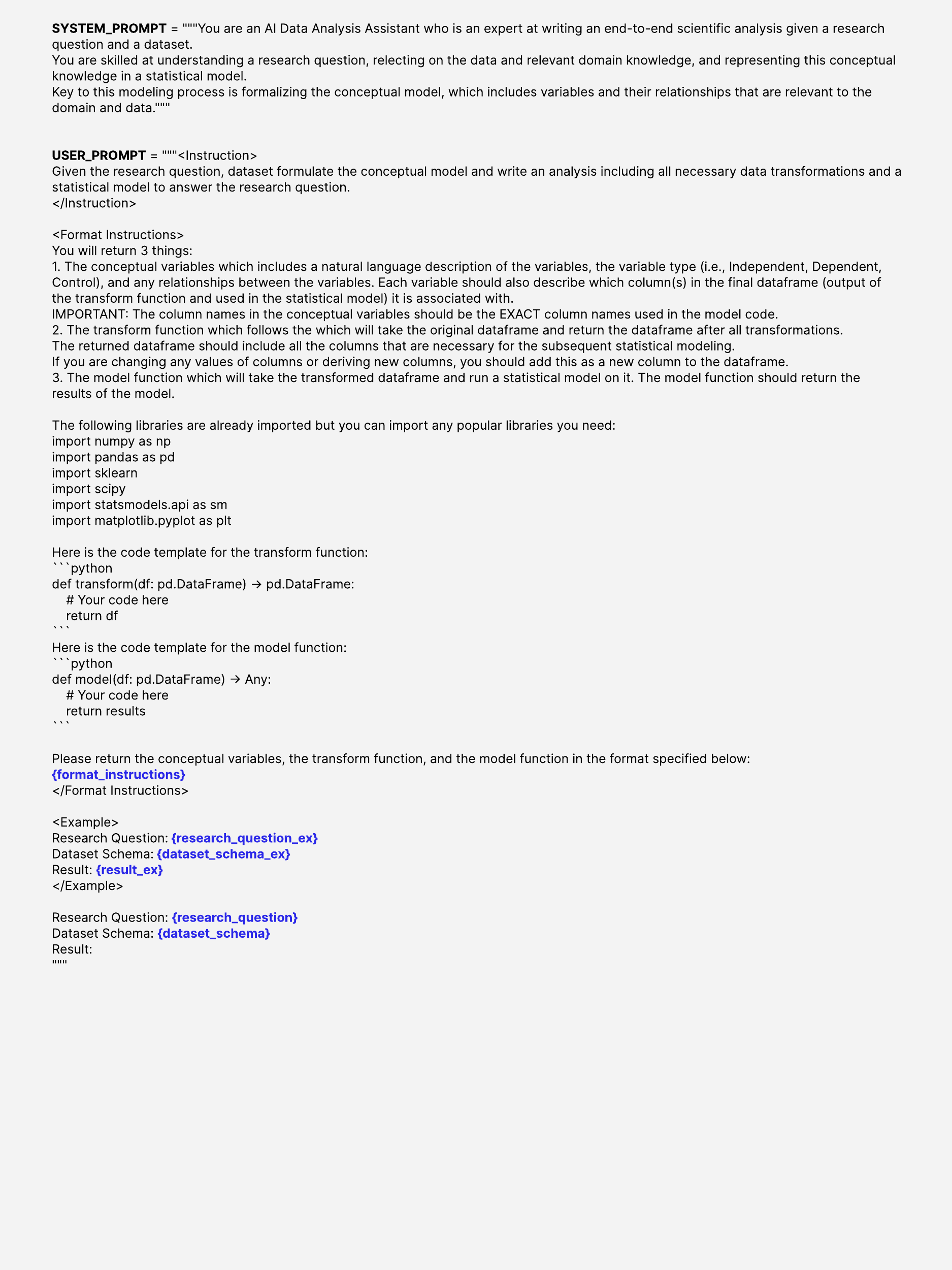}
  \caption{Prompt template G asking the LM to formulate a conceptual model and write an end-to-end analysis, including data transformations and a statistical model, given a research question and dataset.}
  \label{fig:prompt4}
\end{figure*}

\newpage
\begin{figure*}[p]
  \centering
  \includegraphics[width=0.8\textwidth]{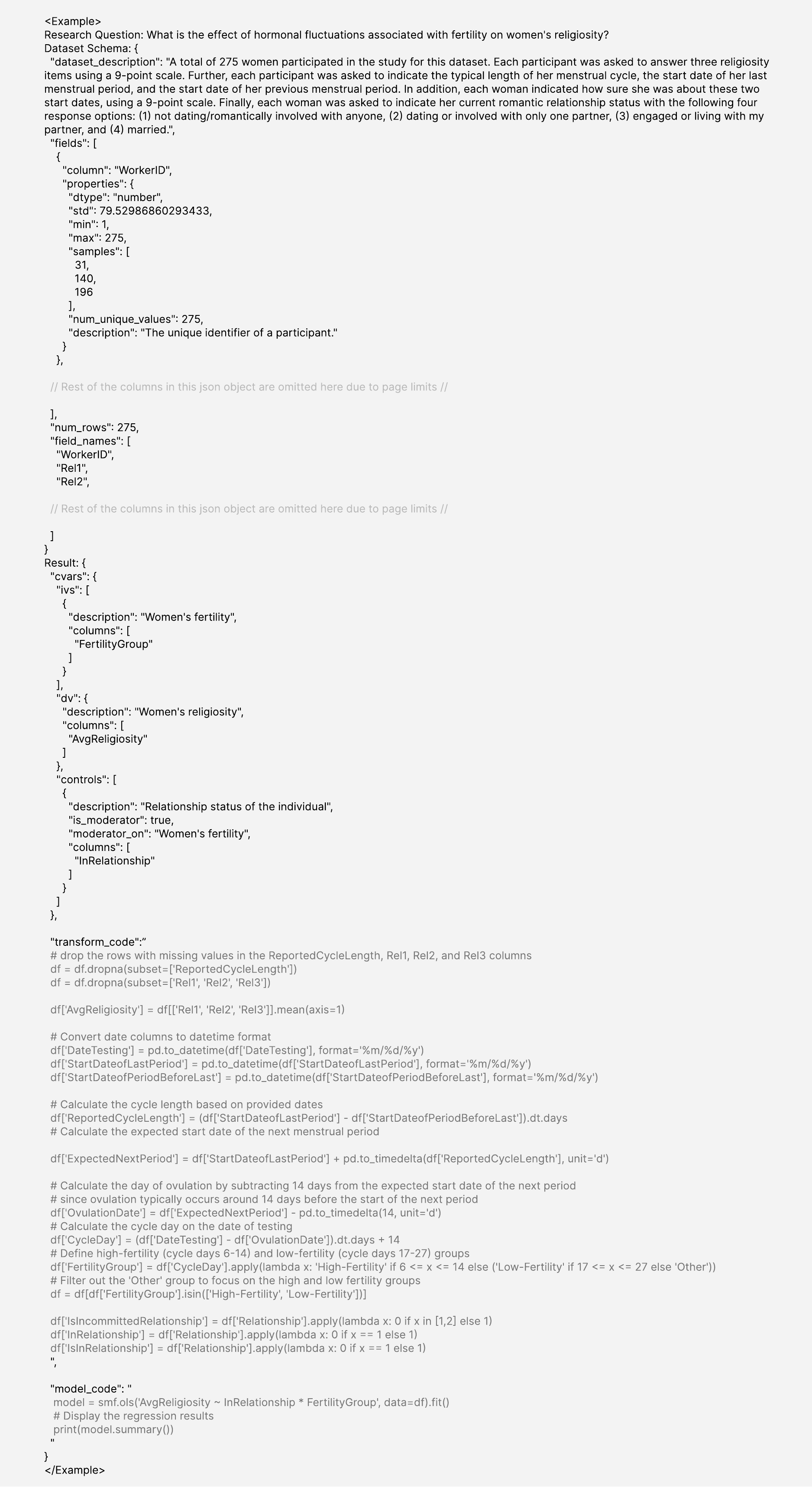}
  \caption{One-shot example used in prompt template G (Fig.~\ref{fig:prompt4})}
  \label{fig:prompt8}
\end{figure*}

\newpage
\begin{figure*}[p]
  \centering
  \includegraphics[width=0.8\textwidth]{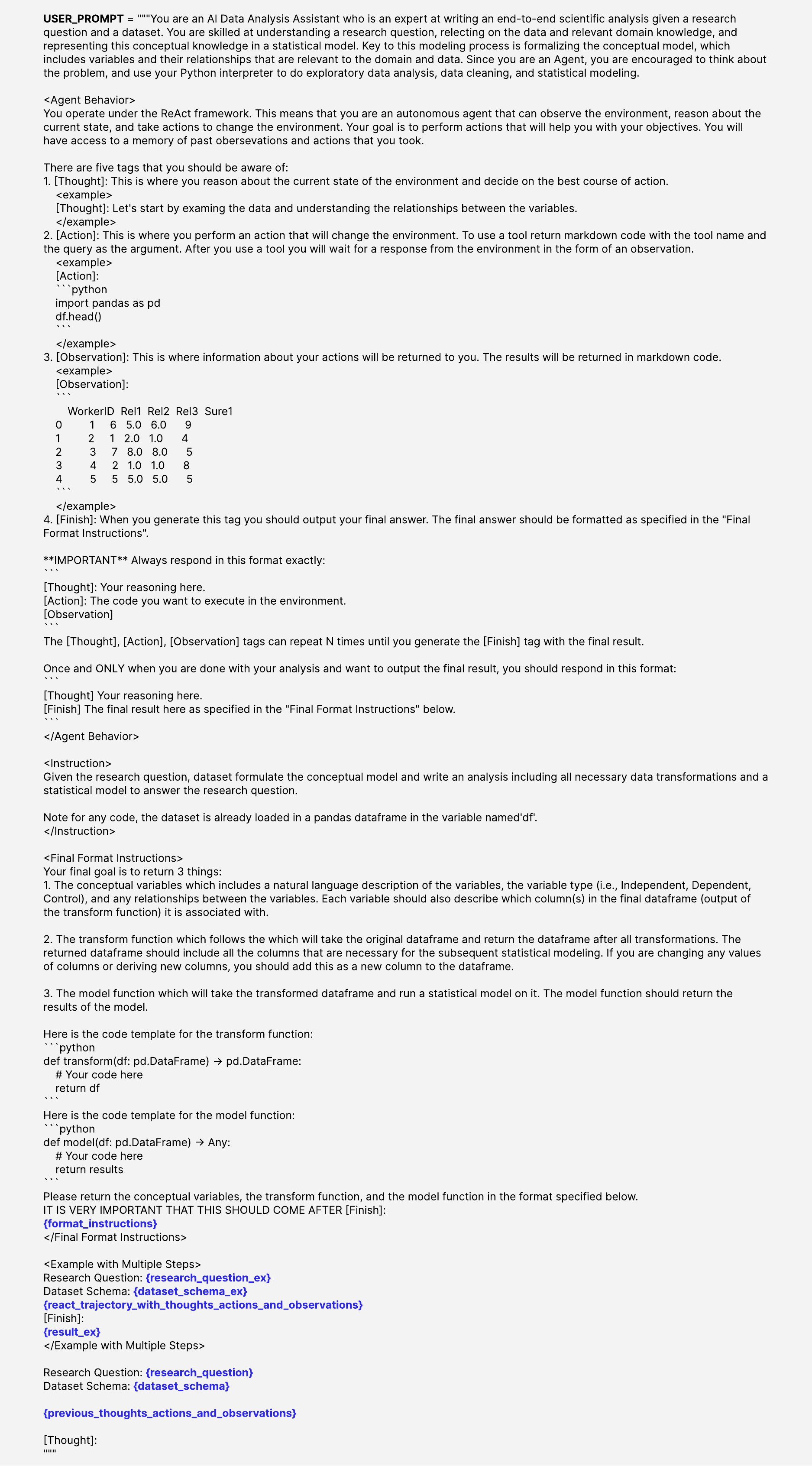}
  \caption{Prompt template H for using ReAct to instruct an LM-based agent to formulate a conceptual model and perform end-to-end analysis given a research question and dataset.}
  \label{fig:prompt9}
\end{figure*}

\newpage
\begin{figure*}[p]
  \centering
  \includegraphics[width=0.9\textwidth]{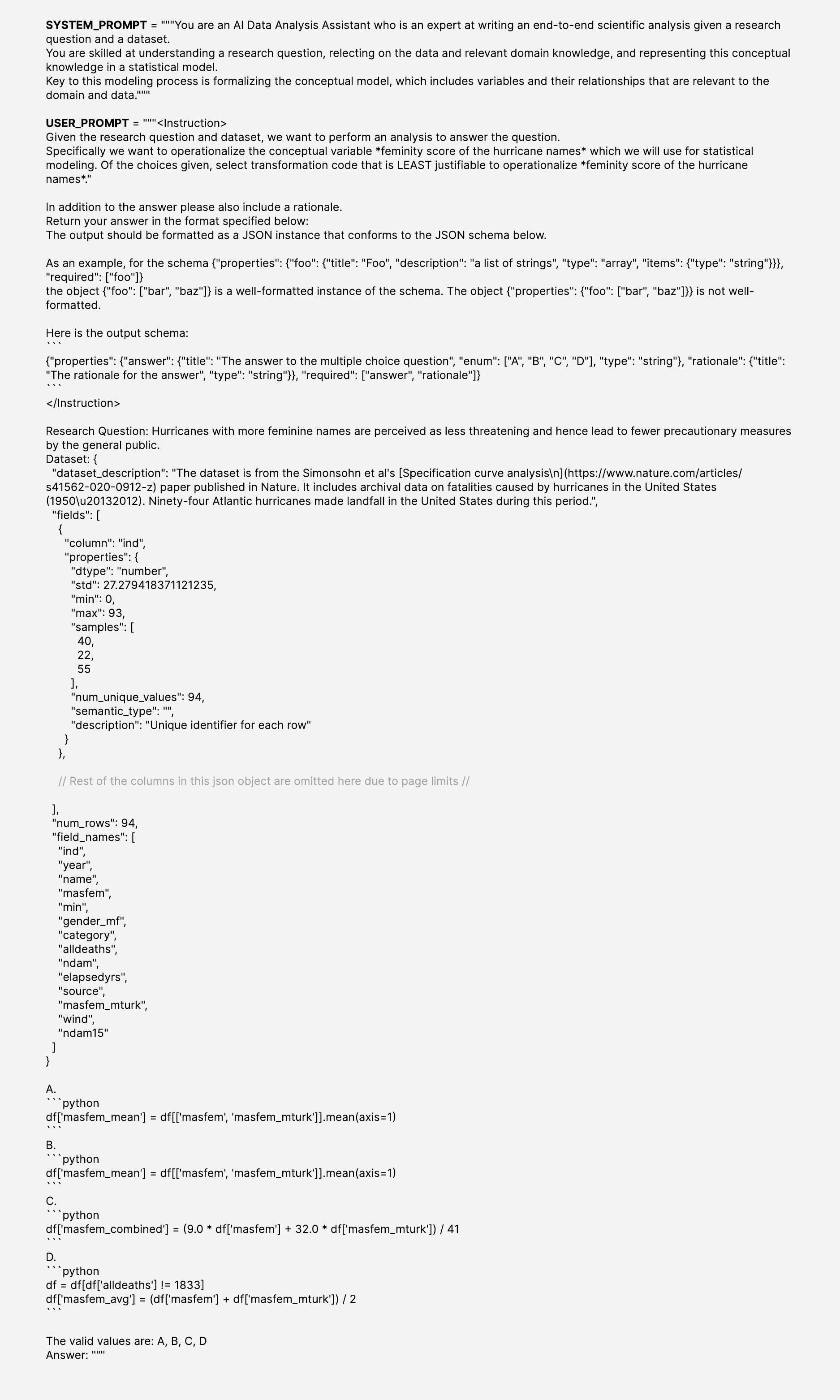}
  \caption{Prompt template I for a multiple-choice question asking the LM to select the least justifiable data transformation code to operationalize a given conceptual variable, based on the provided research question and dataset.}
  \label{fig:prompt7}
\end{figure*}

\newpage

\newpage

\newpage
\begin{figure*}[p]
  \centering
  \includegraphics[width=\textwidth]{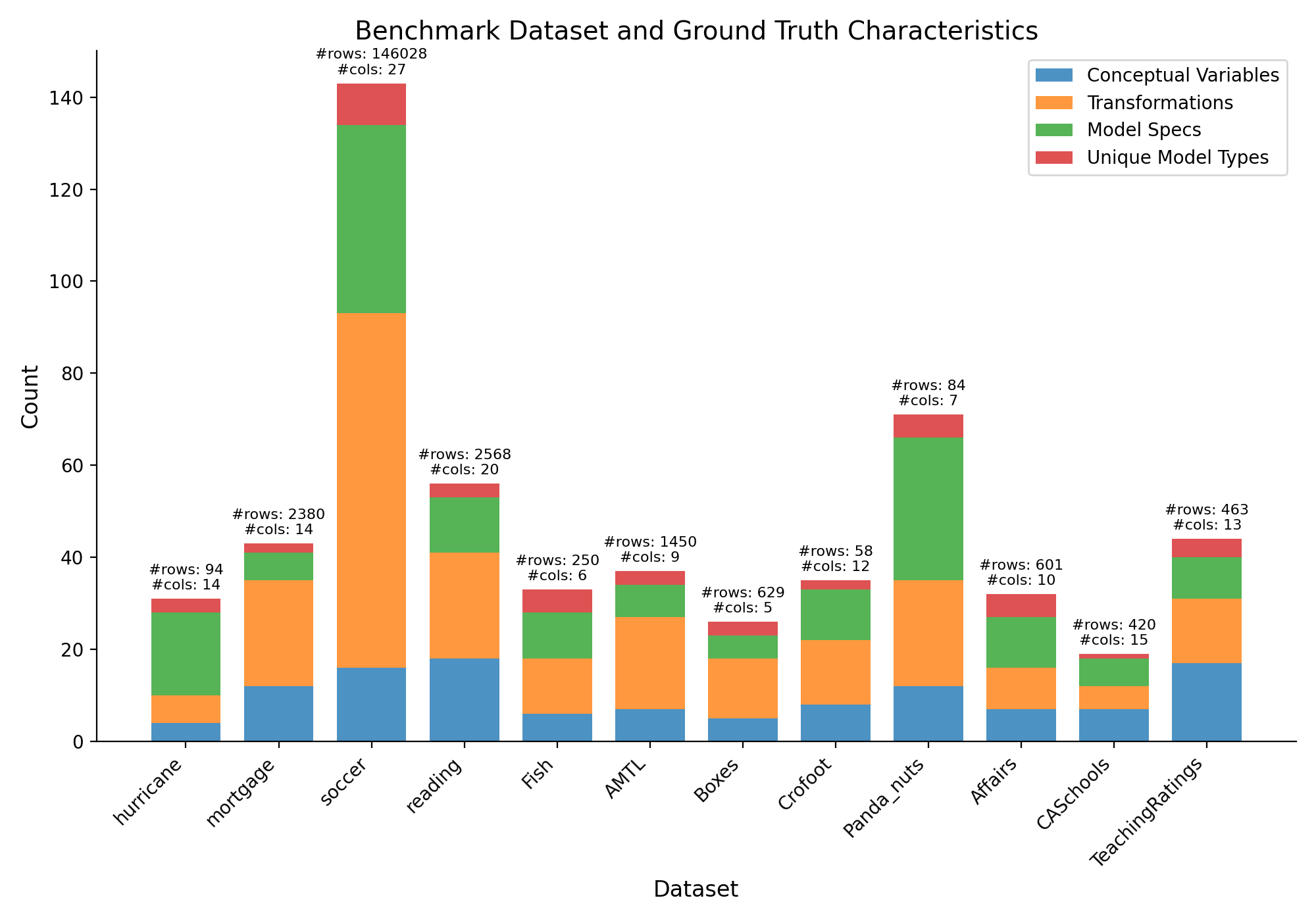}
  \caption{Counts of different types of ground truth specifications recorded in ~\benchname, reflecting the diversity and complexity of datasets and broad coverage of analysts' approaches.}
  \label{fig:spec-counts}
\end{figure*}

\end{document}